\documentclass[default,10pt,colorlinks=true]{sn-jnl}
\usepackage{geometry}
\usepackage[table]{xcolor} 
\usepackage{graphicx}%
\usepackage{multirow}%
\usepackage{amsmath,amssymb,amsfonts}%
\usepackage{amsthm}%
\usepackage{mathrsfs}%
\usepackage[title]{appendix}%
\usepackage{xcolor}%
\usepackage{textcomp}%
\usepackage{manyfoot}%
\usepackage{adjustbox}
\usepackage{booktabs}%
\usepackage{ulem}
\usepackage{algorithm}%
\usepackage{algorithmicx}%
\usepackage{algpseudocode}%
\usepackage{listings}%
\usepackage{makecell}
\usepackage{bbding}
\usepackage{caption}
\newcommand{\revise}[1]{{\color{black}#1}}
\definecolor{colorTHeader}{RGB}{34, 57, 100}
\definecolor{colorTOdd}{RGB}{244, 248, 255}
\definecolor{colorTEven}{RGB}{220, 231, 250}
\theoremstyle{thmstyleone}%
\theoremstyle{thmstyletwo}%

\theoremstyle{thmstylethree}%

\begin{document}
	
	\title[==]{		
ROFI: A Deep Learning-Based Ophthalmic Sign-Preserving and Reversible Patient Face Anonymizer
	}
	

	\author[1,2]{ \fnm{Yuan} \sur{Tian}}
	\equalcont{ These authors contributed equally to this work.}
	\author[1]{ \fnm{Min} \sur{Zhou}}
	\equalcont{ These authors contributed equally to this work.}
	\author[3]{ \fnm{Yitong} \sur{Chen}}
	\equalcont{ These authors contributed equally to this work.}
	\author[1]{ \fnm{Fang} \sur{Li}}
	\author[1]{ \fnm{Lingzi} \sur{Qi}}
	\author[3]{ \fnm{Shuo} \sur{Wang}}
	\author[1]{ \fnm{Xieyang} \sur{Xu}}
	\author[1]{ \fnm{Yu} \sur{Yu}}
	\author[1]{ \fnm{Shiqiong} \sur{Xu}}
	\author[1]{ \fnm{Chaoyu} \sur{Lei}}
	\author[2]{ \fnm{Yankai} \sur{Jiang}}
	\author[2]{ \fnm{Rongzhao} \sur{Zhang}}
	\author[4]{ \fnm{Jia} \sur{Tan}}
	\author[5]{ \fnm{Li} \sur{Wu}}
	\author[6]{ \fnm{Hong} \sur{Chen}}
	\author[7]{\fnm{Xiaowei} \sur{Liu}}	
	\author[8]{\fnm{Wei} \sur{Lu}}
	\author[1]{ \fnm{Lin} \sur{Li}}
	\author[1]{\fnm{Huifang} \sur{Zhou}}
	\author[1]{ \fnm{Xuefei} \sur{Song} \Envelope}\email{songxuefei@shsmu.edu.cn}
	\author[3]{ \fnm{Guangtao} \sur{Zhai} \Envelope}\email{zhaiguangtao@sjtu.edu.cn}
	\author[1]{ \fnm{Xianqun} \sur{Fan} \Envelope}\email{fanxq@sjtu.edu.cn}

	\affil[1]{ Department of Ophthalmology, Shanghai Ninth People's Hospital, State Key Laboratory of Eye Health, Shanghai Key Laboratory of Orbital Diseases and Ocular Oncology, and Center for Basic Medical Research and Innovation in Visual System Diseases of Ministry of Education, Shanghai Jiao Tong University School of Medicine, Shanghai, 200025, China}
	
	\affil[2]{ Shanghai Artificial Intelligence Laboratory, Shanghai, 200000, China}

	\affil[3]{ Institute of Image Communication and Network Engineering, Shanghai Jiao Tong University, Shanghai, 201100, China}
	
	\affil[4]{ Eye Center of Xiangya Hospital, Central South University, Hunan Key Laboratory of Ophthalmology, and National Clinical Research Center for Geriatric Disorders, Xiangya Hospital, Central South University, Hunan, 410008, China}
	\affil[5]{ Department of Ophthalmology, Renmin Hospital of Wuhan University, Hubei, 430060, China}
	\affil[6]{ Department of Ophthalmology of Union Hospital, Medical College, Huazhong University of Science and Technology, Hubei, 430022, China}
	
	\affil[7]{ Department of Ophthalmology, Peking Union Medical College Hospital, Chinese Academy of Medical Science \& Peking Union Medical College, Beijing, 100730, China}
	
	\affil[8]{ The Second Hospital of Dalian Medical University, Dalian, 116023, China}

	\maketitle

\section*{Abstract}
Patient face images provide a convenient mean for evaluating eye diseases, while also raising privacy concerns. Here, we introduce ROFI, a deep learning-based privacy protection framework for ophthalmology. Using weakly supervised learning and neural identity translation, ROFI anonymizes facial features while retaining disease features (over 98\% accuracy, $\kappa > 0.90$). It achieves 100\% diagnostic sensitivity and high agreement ($\kappa > 0.90$) across eleven eye diseases in three cohorts, anonymizing over 95\% of images. ROFI works with AI systems, maintaining original diagnoses ($\kappa  > 0.80$), and supports secure image reversal (over 98\% similarity), enabling audits and long-term care. These results show ROFI's effectiveness of protecting patient privacy in the digital medicine era.

	\section{Introduction}
	
	Image data is crucial for clinical diagnosis~\cite{eisenberg2010clinical,schoser2007ocular,dhawan2011medical}, medical research~\cite{clark2013cancer,mcauliffe2001medical}, as well as artificial intelligence (AI)-aided disease diagnosis~\cite{esteva2017dermatologist,mckinney2020international,shiraishi2011computer,huang2021artificial,zhang2024challenges,2020Automated,Pooja2020An,2023Chapter,2021An,2022Deep}.
	Medical images account for over 90\% of all medical data~\cite{zhao2023multi} and grow by more than 30\% annually~\cite{kiryati2021dataset}.
However, the collection of personal medical images also increases the risk of privacy breaches~\cite{wang2023identifying,ziller2024reconciling,holub2023privacy,mittal2024responsible,heidt2024intellectual,ballhausen2024privacy,wan2022sociotechnical,steeg2024re,tian2025towards}.
Thus, the development of anonymization methods~\cite{muschelli2019recommendations,moore2015identification,newhauser2014anonymization,robinson2014beyond,rodriguez2010open,tian2025medical,tian2025semantics} is increasingly critical, aiming to remove the personal identifiable information from the images.
	Among all medical image types, the facial images draw particular attentions~\cite{newton2005preserving,padilla2015visual,2020Multimodal}, as it includes rich biometric identifying information, and is widely adopted in access control~\cite{yu2024convenient,lee2024user}.

In ophthalmology, facial images play a more prominent role than many other medical specialties~\cite{khumdat2013development,jung2019strabismus,de2021strabismus,zheng2021detection,reid2019artificial,leong2022artificial,shu2024artificial}, particularly for diagnosing external eye conditions like {strabismus, ptosis, and thyroid eye disease} (TED). These eye diseases manifest through visible signs within the periocular areas and the related facial tissues~\cite{graham1974epidemiology,wilson1983adverse,patel2002ocular,Gordana2010Ocular}.
Traditional anonymization techniques, such as cropping out the eye~\cite{yang2024privacy}, blurring~\cite{gedraite2011investigation} or applying mosaics~\cite{battiato2006survey} to faces, are insufficient, since advanced face recognition systems~\cite{deng2019arcface,zhang2019adacos,wang2018cosface} can still identify individuals from these altered images~\cite{ding2017trunk,li2019low,ahmed2018lbph}.
Artificial Intelligence Generated Content (AIGC)-based methods~\cite{tripathy2021facegan,shoshan2021gan,he2019attgan,rombach2022high,podell2023sdxl,kim2022diffusionclip} could further distort the identity, but at the cost of obscuring local details critical for diagnosis.
Face swapping techniques~\cite{korshunova2017fast,chen2020simswap,zhu2021one,xu2022region,naruniec2020high,chadha2021deepfake} replace original facial features with those of another individual, such as a celebrity's~\cite{karras2017progressive}.
Similarly, face de-identification methods~\cite{li2019anonymousnet,gu2020password,li2023riddle,yang2024g,cai2024disguise,wen2023divide,cao2021personalized} transform the face into a totally virtual one. While protecting privacy, the above approaches also heavily alter the clinical signs necessary for eye disease diagnosis.

Therefore, it is urgent to develop ophthalmic sign-preserving facial anonymization methods.
Indeed, there are few preliminary efforts~\cite{yang2022digital,zhu2024facemotionpreserve,cai2024disguise} on this, which adopt hand-crafted clinical features such as facial morphology~\cite{paysan20093d,li2017learning,feng2021learning} parameters and facial keypoints~\cite{sun2013deep,rashid2017interspecies} to partially represent the disease signs.
These hand-crafted features are enough for diagnosing a limited spectrum of diseases such as ptosis~\cite{marenco2017clinical}, but are incapable of handling many other complex conditions.
For instance, conjunctival melanoma~\cite{seregard1998conjunctival} manifests with diverse conjunctival pigmentation and lesion patterns involving multiple regions.
Squamous Cell Carcinoma~\cite{marks1996squamous} presents with highly-variable red patches or nodules on the eyelid or ocular surface.
Additionally, the above methods are not reversible, limiting their ability to retrieve personal medical records for decision-making~\cite{rothman2012future} and medical audits~\cite{smith1990medical,act1996health}.

\begin{figure*}[!thbp]
		\centering
		\includegraphics[width=0.95 \linewidth]{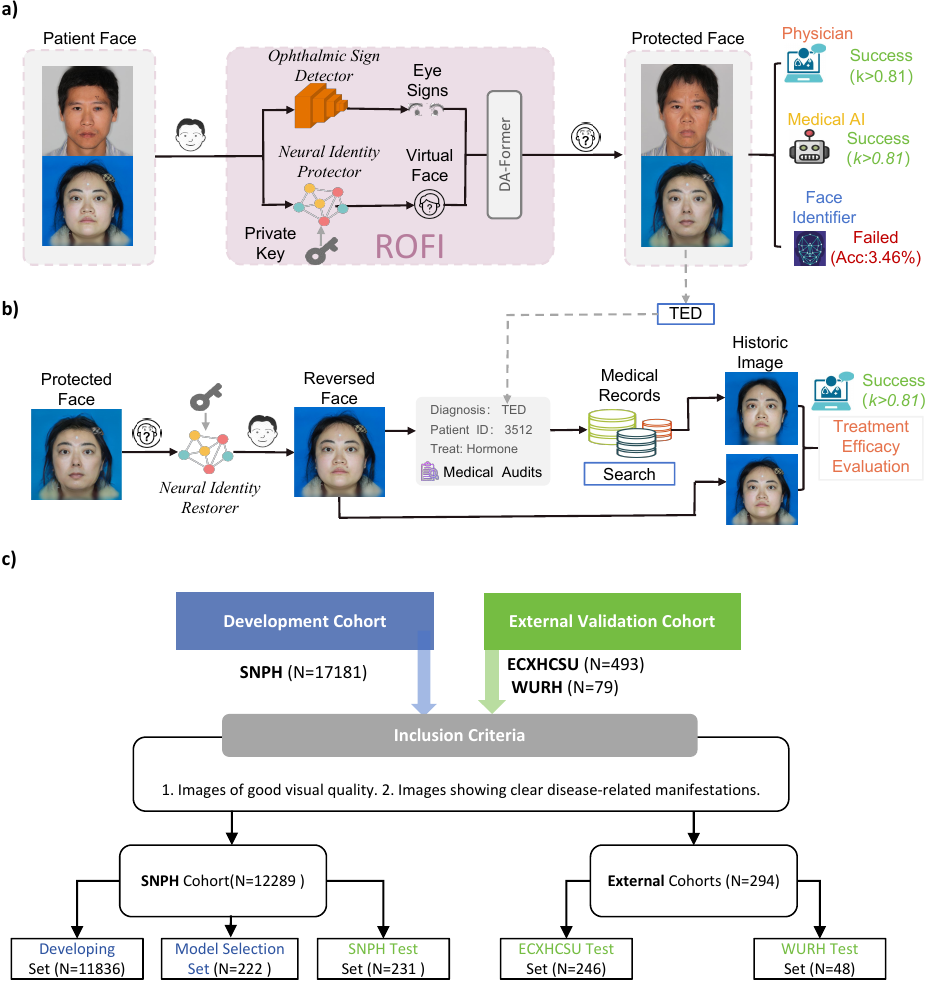}
		\caption{
			\textbf{Study Overview.}
			\textbf{a)} ROFI safeguards patient privacy while preserving eye disease-related signs, by introducing two data-driven deep neural networks: the Ophthalmic Sign Detector and the Neural Identity Protector. The Ophthalmic Sign Detector identifies and retains ophthalmic signs of the original image, while the Neural Identity Protector alters bio-identifying facial information. These features are combined and refined by the DA-Former to produce final high-quality images that are both privacy-preserving and clinically usable.
			The protected images can be readily diagnosed by human physicians and medical AIs. Diagnostic consistency is measured using Cohen's Kappa value ($\kappa$), where $k > 0.81$ indicates remarkable agreement.
			\textbf{b)} 
			ROFI supports confidential reversibility with another deep network Neural Identity Restorer, enabling the accurate reconstruction of original images, using a private key established during the protection process. The reversibility is critical for medical audits, ensuring that patient information can be traced-back and accurately recorded.
			Moreover, this facilitates personalized medical record retrieval, supporting longitudinal evaluations, such as the assessment of {thyroid eye disease} (TED) therapy outcomes.
			\textbf{c)} Cohort building. The developing set is used for developing the ROFI models, and a model-selection set is adopted for selecting the best model. One internal and two external validation sets are built to evaluate the selected ROFI model.
			SNPH: Shanghai Ninth People's Hospital; ECXHCSU: Eye Center of Xiangya Hospital of Central South University; WURH: Renmin Hospital of Wuhan University.
			Icons are from https://uxwing.com/ and https://www.biorender.com/.
			Created with BioRender.com.
		}
		\label{fig:paradigm_compare}
		%	\vspace{-0.9cm}
	\end{figure*}

In this article, we introduce ROFI (Figure~\ref{fig:paradigm_compare}a,b), the first ophthalmic sign-preserving and reversible facial anonymization approach. ROFI is achieved through two key designs:
 (1) {Data-Driven Ophthalmic Sign Detection}: Using weakly-supervised learning~\cite{hsu2019weakly,lu2021data} techniques, given a large-scale set of patient faces annotated with binary labels (whether or not they have eye disease), a neural network automatically separates disease-related signs from facial images. This achieves autonomous sign detection with affordable data annotation cost.
(2) {Reversible Neural Identity Translation}: Leveraging the flexible feature translation capability of the Transformers~\cite{vaswani2017attention}, we build a pair of Neural Identity Protector and Neural Identity Restorer models to achieve reversible transformation of facial identities. {The procedures are governed by a customized privacy key.}
This enables accurate reconstruction of original images when being authorized.
Compared to previous approaches~\cite{chen2020simswap,yang2024g,yang2022digital}, our approach is uniquely usable throughout the entire medical process.
{During early screening of eye diseases using facial images, ROFI largely alleviates the patient privacy issue.}
In subsequent rigorous examinations, it reconstructs the original images to refer to the history medical records.

To verify the effectiveness of ROFI, we conducted a comprehensive study across three clinical centers (Figure~\ref{fig:paradigm_compare}c) enrolling 17,181 patients from Shanghai Ninth People's Hospital (SNPH), 493 patients from Eye Center of Xiangya Hospital of Central South University (ECXHCSU), and 79 patients from Renmin Hospital of Wuhan University (RHWU). We evaluated the clinical usability of ROFI in two key aspects: 1) the completeness of ophthalmic signs, and 2) the effectiveness for screening eleven ophthalmic diseases with external manifestations. Then, we explored the usability of ROFI-anonymized images for medical AI models.
Further, we assessed the facial anonymization capability with modern face recognition systems.
Finally, we evaluated the similarity between the reversibly reconstructed images and the originals, and utilized the reconstructed images to retrieve historical medical records, assessing the efficacy of hormone therapy for {thyroid eye disease} (TED).

	\section{Results}

	\subsection{Cohort Building}
	For the development and evaluation of ROFI, we included 17181 patients who had undergone ophthalmologic examinations and had facial images captured at three hospitals in China between January 1, 2018, and December 25, 2023. All patients involved in this study have agreed to participate in the project either by themselves or via their legal guidance.
	The derivation cohort comprised patients from Shanghai Ninth People's Hospital (SNPH), which served as the primary site for training and internal validation.
	Patients were included in the study if they met the following criteria:
	(1) Had at least one facial image available that was taken within the study period.
	(2) Were diagnosed with ophthalmic diseases characterized by external manifestations.
	(3) Provided images that contained sufficient quality and clarity for ophthalmic sign detection.
	Patients were excluded if:
	(1) Their facial images were of insufficient quality, such as being too blurry or having significant occlusions that could interfere with the identification of ophthalmic signs.
	This resulted in the final SNPH cohort comprising 12,289 patients. We randomly divided the SNPH cohort into three subsets: 11,836 for developing, 222 for model-selection, and 231 for internal validation.
	For external validation, two additional cohorts were established, ECXHCSU and RHWU, with 246 and 48 patients, respectively. These patients were enrolled using the same criteria as the SNPH cohort. The external test sets aim to assess the ROFI model's generalization across different clinical settings.
	
	Given the extensive annotation effort required for disease-related region mask labeling, developing set images were annotated only with binary labels indicating patient health status. Weakly-supervised learning techniques were then applied to autonomously localize disease signs, thereby reducing the physicians' workload.
	For the validation sets, images were annotated with the corresponding ophthalmic disease name.
	The study was approved by the Institutional Review Boards of Shanghai Ninth People's Hospital (Approval No.SH9H-2022-T380-1), Eye Center of Xiangya Hospital of Central South University (Approval No.202407131), and Renmin Hospital of Wuhan University (Approval No.WDRY2024-K238).
	All procedures involving human participants in this study were conducted in accordance with the ethical standards of the institutional research committee and the latest version of the Declaration of Helsinki~\cite{bierer2024declaration}. Informed consent was obtained from all individual participants included in the study. For participants who were minors or otherwise unable to provide consent independently, written informed consent was obtained from their parents or legally authorized representatives. Participation in the prospective study at the ROFI Program was voluntary, and all individuals (or their legal representatives) were fully informed about the nature, purpose, potential risks, and benefits of the study prior to enrollment. Participants were assured of their right to withdraw from the study at any time without any consequences to their care or treatment.

	\subsection{Model Development}
	We have developed the ROFI framework, a reversible patient face privacy protection approach that particularly preserves ophthalmic disease signs (Figure~\ref{fig:paradigm_compare}a). Unlike previous methods that either lack ophthalmic sign modeling~\cite{cao2021personalized,li2023riddle,yang2024g} or rely solely on hand-crafted sign features~\cite{yang2022digital,zhu2024facemotionpreserve}, we introduced a learnable ophthalmic sign detector, which autonomously mines eye disease-related sign features from vast patient facial images in a weakly-supervised manner. The detected sign features are then used to substitute the ophthalmic features that are altered during the privacy protection procedure.
	To amend any artifacts caused by the feature substitution process, we employ a Transformer~\cite{vaswani2017attention}-based feature enhancement network, called DA-Former to refine these features, finally producing the high-quality anonymous patient faces.
	
	The sign detector is trained on the patient facial images of developing set, which is labeled with binary health state classifications (healthy or not), with a weakly-supervised region-score-max learning strategy.
	Subsequently, the neural identity translator, and DA Transformer are jointly optimized, ensuring the generated images are well-protected, highly-realistic, and highly clinically usable.
	The model details and implementation are described in the Method section.

	\subsection{ROFI Well Preserved Eye Signs}
	Eye sign evaluation is the fundamental of diagnosing ophthalmic diseases. For instance, the typical symptom of ptosis is the drooping of the upper eyelid~\cite{diaz2018ocular}. Ocular melanoma is typically associated with the conjunctiva~\cite{wong2014management}, and some malignant tumors~\cite{stannard2013radiotherapy} can lead to pupil abnormalities. In this section, we have chosen the following typical eye signs that are relevant to most eye diseases: eyelid shape, iris shape, conjunctival disorder (Conj-Dis)~\cite{azari2013conjunctivitis}, lacrimal apparatus disorder (LA-Dis)~\cite{jones1961anatomical}, eyelid disorder (Eyelid-Dis)~\cite{yin2015eyelid}, and iris disorder (Iris-Dis)~\cite{aslam2009iris}. The former two shapes were quantified using eyelid/iris keypoints, which were automatically detected by ocular keypoint detection models~\cite{2009Dlib,mediapipe_face_mesh}. The latter four disorders were assessed manually by ophthalmologists, with each image being reviewed by three specialists. The final diagnosis was determined by majority voting.
	{More explanation of the above signs are provided in the Supplementary Information.}

	To better demonstrate the superiority of ROFI, we compared it with five representative facial privacy protection methods: the conventional approach that applies mosaic to the image (Mosaic)~\cite{battiato2006survey}, using state-of-the-art AI generation model~\cite{rombach2022high} to regenerate facial content (AIGC), famous face swapping software SimSwap~\cite{chen2020simswap} (Face Swap), the method~\cite{yang2022digital} specifically designed for ophthalmic use (Digital Mask), and the state-of-the-art face de-identification method~\cite{yang2024g} (G2Face).
	{The comparison of these methods is given in the supplementary Table 15.}

	ROFI accurately preserved ophthalmic signs across both internal and external validation sets. For example, on SNPH set, it exhibited the lowest keypoint error rate (Figure~\ref{fig:quan_results}b and Supplementary Table 1), with an average of 1.53\%/1.64\% for eyelid/iris keypoint errors, much lower than Mosaic (5.51\%/12.36\%), AIGC (3.19\%/3.79\%), Face Swap (2.35\%/4.16\%), Digital Mask (8.25\%/6.84\%), and G2Face (2.70\%/3.18\%).
	On the external validation sets, it still achieved very small eyelid/iris keypoint errors, 1.38\%/1.37\% and 1.68\%/2.07\% on ECXHCSU and WUPH, respectively (Supplementary Figure 1 and Supplementary Table 1).
	Besides, ROFI also outperforms other approaches for per-disease comparison (Supplementary Figure 2 and Supplementary Figure 3).
	The above results demonstrated ROFI is highly capable of maintaining ocular morphological features for various eye diseases across different cohorts.
	
	Regarding disorder signs, images processed by ROFI showed remarkable consistency (Cohen's $k \geq 0.81$) with original images across all three validation sets (Figure~\ref{fig:quan_results}d and Supplementary Table 3). On the internal SNPH set, taking the Conj-Dis sign as an example, ROFI achieved a $\kappa$ value of 0.9223 (95\% CI: 0.8691-0.9755), surpassing Mosaic ($k=$0.3299, 95\% CI: 0.2079-0.4518), AIGC ($k=$0.2399, 95\% CI: 0.1260-0.3538), Face Swap ($k=$0.5208, 95\% CI: 0.4176-0.6239), Digital Mask ($k=$0.0955, 95\% CI: -0.0250-0.2162), and G2Face ($k=$0.4489, 95\% CI: 0.3406-0.5571). ROFI significantly outperformed the state-of-the-art deep facial privacy protection method G2Face for all ophthalmic signs on all validation sets, which is reflected by the very small $P$ values ($P < 0.001$ for three out of four signs on SNPH and all four signs on ECXHCSU and WUPH). ROFI demonstrated much higher eye structure similarity to the original images compared to other methods. For instance, on ECXHCSU, ROFI achieved 94.09\% similarity, outperforming Mosaic (55.55\%), AIGC (53.74\%), Face Swap (74.15\%), Digital Mask (41.37\%), and G2Face (68.38\%) (Supplementary Table 2).
	These findings suggest that ROFI is superior in preserving disease-related ophthalmic signs during face anonymization.
	
	 Qualitative comparisons (Figure~\ref{fig:quan_results}e) revealed that only our ROFI generated high-quality anonymous facial images with well-preserved ophthalmic signs. In contrast, traditional method Mosaic obscured all ophthalmic signs. AIGC generated high-quality facial images, but it also introduced random signs, making clinical application infeasible. Face Swap retained some basic ophthalmic attributes, such as iris color, but also introducing other new artifacts that misleads diagnostic results.
	 G2Face had the same artifact generation problem, even more severe.
Digital Mask preserved only hand-crafted eyelid/iris shapes, neglecting other important signs necessary for Corneal Leukoma.

	\subsection{ROFI Enabled Accurate Clinical Diagnosis}
	To evaluate the effectiveness of our ROFI in disease diagnosis, a clinical comparison was conducted. The evaluation of privacy-removed images was conducted as two phases: health state assessment (Yes-or-No) and initial disease diagnosis (Diagnosis) (Figure~\ref{fig:trail_clinical}a).
	In the first phase, three physicians were tasked with classifying each image as healthy or unhealthy. The objective here was to detect all potential positive cases possible, so that individuals showing signs of disease could proceed to a medical facility for further examination.
	For the second phase, which focuses on disease diagnosis, the same three physicians were consulted to categorize each diseased image. Given that this study does not incorporate additional medical materials such as fundus images~\cite{bernardes2011digital}, directly diagnosing diseases solely from facial images would be challenging and impractical. Therefore, the diagnosis process was simplified into a selection task, where disease names were chosen from a predefined list.

	For the health state assessment (Yes-or-No), our ROFI achieved sensitivities of 100\% (95\% CI: 100\%-100\%), 100\% (95\% CI: 100\%-100\%), and 100\%(95\% CI: 100\%-100\%) on the SNPH, ECXHCSU, and WUPH validation sets, respectively (Figure~\ref{fig:trail_clinical}b and Supplementary Table 4). Given that sensitivity measures the ratio of detected positive samples to all positive samples, our approach detected all patients with eye disease and reminded them to go to the hospital, while all other approaches failed.
	For example, on WUPH, the sensitivity achieved by our approach (100\%, 95\% CI: 100\%-100\%), significantly outperformed Mosaic (85.00\%, 95\% CI: 74.36\%-95.00\%), AIGC (77.50\%, 95\% CI: 64.28\%-89.74\%), Face Swap (97.50\%, 95\% CI: 91.89\%-100.00\%), Digital Mask (85.00\%, 95\% CI: 72.97\%-95.12\%), and G2Face (85.00\%, 95\% CI: 74.27\%-95.12\%).
	All other methods missed a considerable number of positive cases, which could potentially delay diagnosis and treatment. In contrast, ROFI timely facilitates medical intervention for individuals suffering from eye diseases.

For disease diagnosis task, we evaluated the medical outcomes of images protected by different methods using Cohen's Kappa ($\kappa$) values. Our method achieved $\kappa \geq 0.81$ across all three validation sets for all eye diseases (Figure~\ref{fig:trail_clinical}c and Supplementary Table 5), indicating excellent clinical agreement between the results obtained from the ROFI-protected and the original images. For instance, on SNPH, ROFI obtained $\kappa$ values of 0.9594 (95\% CI: 0.9033-1.0154) for BCC, 0.9046 (95\% CI: 0.7735-1.0357) for CM, 0.8732 (95\% CI: 0.7318-1.0147) for CL, 1.0 (95\% CI: 1.0-1.0) for SCC, and similar high values for {strabismus, ptosis}, and TED.
Given that a $\kappa$ value exceeding 0.81 signifies remarkable agreement, ROFI can be effectively deployed for remote clinical diagnosis application, without obviously compromising clinical outcomes.

In contrast, Mosaic and AIGC performed poorly across all diseases due to the removal of critical disease signs during processing. Digital Mask showed good results for conditions like TED and ptosis, which can be assessed with hand-crafted features such as eyelid shape, but it failed with other diseases such as BCC and CM, achieving poor $\kappa$ values of $0.0712$ (95\% CI: -0.0986-0.2409) and $0.0993$ (95\% CI: -0.1400-0.3386) for these two diseases on ECXHCSU, respectively. This was far below our approach, which achieved $\kappa = 0.9692$ (95\% CI: 0.9091-1.0294) and $\kappa = 1.0$ (95\% CI: 1.0-1.0) for the above two conditions on ECXHCSU.
Face Swap and G2Face showed better performance than Mosaic, AIGC, and Digital Mask, but still did not reach $\kappa > 0.81$ for any eye disease, suggesting their unsuitability for clinical use.
These findings highlighted ROFI as the first clinically-friendly facial image privacy protection method that can be deployed to both immunologic (TED), congenital (e.g., {ptosis and microblepharon}), corneal (CL), and tumor (e.g., BCC, SCC, and CM) eye diseases.

Besides, ROFI achieved a Matthews Correlation Coefficient (MCC) value of 0.9450 with 95\% CI: 0.8684-1.0000 on WUPH, which is also much better than G2Face 0.5157 with 95\% CI: 0.3705-0.6895, Digital Mask 0.4960 with 95\% CI: 0.3390-0.6422, Face Swap 0.6501 with 95\% CI: 0.5142-0.8318, AIGC 0.1513 with 95\% CI: 0.0016-0.3394, and Mosaic 0.1604 with 95\% CI: 0.0449-0.3442 (Supplementary Table 6 and Supplementary Figure 4), further indicating the superior clinical outcome of ROFI.

The confusion matrix highlighted that previous methods are effective for only a subset of eye diseases (Figure~\ref{fig:trail_clinical}d, Supplementary Figure 5 and Supplementary Figure 6). For example, on WUPH, Face Swap and G2Face excelled with BCC and ptosis but faltered on TED, while Digital Mask performed well on TED and ptosis but failed on BCC. In contrast, our approach maintained well performance across all disease categories.

{Additionally, we compare ROFI with other approaches on two other tasks, namely, facial landmark detection and tumor segmentation.
	For the facial landmark detection, we evaluate with the large-scale CelebA-HQ dataset~\cite{karras2017progressive}.
	We adopt the MTCNN~\cite{zhang2016joint} to automatically detect the landmarks of the original and the protected images, and quantify the performance with mean squared error.
	For the tumor segmentation task, we engaged physicians to annotate the tumor regions for Basal Cell Carcinoma (BCC) instances in SNPH set.
	The Dice coefficients~\cite{bertels2019optimizing} between the tumor regions from the original and the protected images by various approaches are calculated and compared.

	As shown in Supplementary Table 7, our method achieves superior accuracy in landmark detection compared to existing techniques.
	Concretely, the error of landmarks detected from the ROFI-protected images is 2.9871, which is much lower than Mosaic (5.6799), AIGC (7.8945), Face Swap (4.8921), Digital Mask (6.3891), and G2Face (3.7130).
	The accurate landmark preservation capability of our approach also implies that it could be potentially deployed to other medical conditions. For instance, precise facial landmark detection is crucial for analyzing facial symmetry, a key clinical indicator in diagnosing conditions such as facial palsy or Bell’s palsy~\cite{roob1999peripheral}.
	To evaluate the capability of our method in fine-grained disease detection, we collaborated with board-certified physicians to annotate the tumor regions in BCC (Basal Cell Carcinoma) instances within the SNPH validation set.
	As shown in Supplementary Table 8, our approach achieves a Dice coefficient~\cite{bertels2019optimizing} of 0.7389, largely outperforming previous methods such as G2Face (0.5782) and Face Swap (0.2783).
	This substantial improvement demonstrates that our method is not only effective for coarse-level disease classification, but also capable of performing fine-grained disease severity assessment.
	}

	\subsection{ROFI was Friendly to Medical AI Models}
	An increasing number of artificial intelligence (AI)~\cite{esteva2017dermatologist,mckinney2020international,shiraishi2011computer,huang2021artificial,zhang2024challenges} models are being developed to predict patient health conditions from medical images, with some now deployed in real-world applications, e.g., video-based disease screening~\cite{li2024performance,tian2019video,tian2024coding,tian2022ean,tian2023clsa,tian2023non,tian2024free,tian2021self,chen2024gaia,tian2025smc++}. It is crucial that protected images by ROFI remain recognizable by AI models.
	To verify this, we divided the SNPH and ECXHCSU datasets into training and test subsets; WUPH was not used due to its too small data size to train an AI model. The models were trained on the training subset and evaluated on test sets processed by different privacy protection methods.
	We developed facial image-based eye disease diagnostic models based on two neural network architectures: the classical convolutional neural network (CNN) model ResNet50~\cite{he2016deep} and a recently-popular Transformer-based model ViT~\cite{dosovitskiy2020image}. Then, we use the trained models to assess the impact of different privacy protection methods on the AI diagnosis performance.
	The details are provided in the Method section.

	Our approach, ROFI, achieved excellent consistency with results obtained from original images, as indicated by the highest Cohen's Kappa values ($\kappa$) among all privacy protection methods.
	For instance, on SNPH using the ResNet50 diagnosis model, ROFI achieved a $\kappa$ value of 0.9077 (95\% CI: 0.8569--0.9586), significantly outperforming the state-of-the-art facial privacy protection method G2Face ($\kappa = 0.7257$, 95\% CI: 0.6454--0.8060) (Figure~\ref{fig:trail_AI_model_auc}a).
	Similarly, with the ViT diagnosis model, ROFI demonstrated significantly higher $\kappa$ values compared to other approaches (Figure~\ref{fig:trail_AI_model_auc}b).

	Our approach achieved an Area Under the Receiver Operating Characteristic curve (AUROC) of 0.9113 (95\% CI: 0.8952-0.9113) on SNPH, when being evaluated using the classic ResNet50, remarkably outperforming Mosaic (AUROC: 0.6102; 95\% CI: 0.5941-0.6262), AIGC (AUROC: 0.7438; 95\% CI: 0.7262-0.7614), Face Swap (AUROC: 0.7983; 95\% CI: 0.7781-0.8186), Digital Mask (AUROC: 0.7853; 95\% CI: 0.7816-0.7891), and G2Face (AUROC: 0.8776; 95\% CI: 0.8604-0.8949) (Figure~\ref{fig:trail_AI_model_auc}c).
	Using the ViT, our method attained AUROCs of 0.9124 (95\% CI: 0.9059-0.9190) on SNPH and 0.7894 (95\% CI: 0.7715-0.8072) on ECXHCSU, significantly outperforming other approaches (Figure~\ref{fig:trail_AI_model_auc}c).

	Furthermore, we compared the ROC curves and per-disease AUROCs of our approach with other approaches (Figure~\ref{fig:trail_AI_model_auc}d, Supplementary Figure 7, Supplementary Figure 8 and Supplementary Figure 9). Our approach exhibited ROC curves more similar to those derived from original image data. For per-disease AUROCs, our method either matched or exceeded the performance of the other two approaches. For instance, for the ViT diagnostic model on SNPH, for BCC, our method achieved an AUROC of 0.950, compared to 0.908 for Face Swap and 0.904 for G2Face. These findings highlighted the high compatibility of our method to medical AI models.
	{We notice that, in some settings, our method slightly outperforms the results obtained with the original images. We conjecture that this improvement arises, because our protected images enhance the representation of eye disease-related features, while reducing individual-specific details. This dual effect likely contributes to more robust diagnostic accuracy.}

	We also trained a stronger ViT model using input images of size 512$\times$512.
	As shown in Supplementary Table 9, increasing the image resolution leads to a noticeable improvement in AI diagnostic performance across all privacy-preserving approaches. Nevertheless, our ROFI still demonstrates a significant advantage, achieving an AUROC of 94.59\%, substantially outperforming the second-best method, G2Face, which achieved 86.75\%.

	\subsection{ROFI Protected Patient Privacy and Enabled Reversibility}
	
	To assess the efficacy of the proposed ROFI technique in protecting patient privacy, we conducted an investigation into its performance on evading face recognition systems. In a typical scenario, a face recognition network extracts facial features from protected images and compares them against ID card photos of patients.
	The patient ID corresponding to the image with the closest feature distance is adopted as the recognition result.
	We conducted the experiments on the SNPH set, as the ID card information is extremely sensitive and difficult to acquire from other external sets.
	
	We evaluated the privacy protection performance using two face recognition algorithms: AdaCos~\cite{zhang2019adacos} and ArcFace~\cite{deng2019arcface}. When using the AdaCos~\cite{zhang2019adacos}, ROFI protected 96.54\% of images, surpassing Mosaic (90.91\%), AIGC (93.08\%), Face Swap (74.90\%), Digital Mask (94.81\%), and G2Face (93.51\%) (Figure~\ref{fig:face_recognition}b).
	With ArcFace~\cite{deng2019arcface}, ROFI maintained strong protection rates (Figure~\ref{fig:face_recognition}b).
	
	Furthermore, cropping the eye region, a traditional method for protecting privacy in ophthalmology~\cite{yang2024privacy}, provides insufficient protection. With the ArcFace face recognition method, this approach only safeguarded 68.40\% and 74.03\% of patients for left and right eyes, respectively, compared to 94.81\% with our method (Figure~\ref{fig:face_recognition}c).
	This finding underscores the urgent need for a novel patient privacy protection method, as the current ``unsafe'' eye-cropping strategy is widely employed in ophthalmic departments for privacy purposes. Our protection model ROFI fills this blank.
	
	To better contextualize ROFI's performance, we compare it against standard pixel-wise~\cite{fan2018image} and feature-wise~\cite{xue2021dp} Differential Privacy (DP) methods. We calibrated the noise level in the DP methods to achieve an identity recognition rate (~3.0\%) comparable to that of ROFI. The results (Supplementary Table 13) show that while providing a similar level of empirical privacy, ROFI maintains a significantly higher diagnostic utility (91.25\% AUROC) compared to both pixel-wise DP (62.32\% AUROC) and feature-wise DP (76.69\% AUROC) on SNPH validation set with ViT-based diagnosis model.
	The above results show the advantage of our architecture in utility-privacy trade-off.
\revise{For a more rigorous evaluation, we adopted a state-of-the-art membership inference attack framework based on shadow modeling~\cite{carlini2022membership, nasr2021adversary}. We used this framework to calibrate the feature-wise DP method to a comparable level of empirical privacy as ROFI, specifically a True Positive Rate (TPR) of approximately 1.4\% at a 1\% False Positive Rate (FPR). The full details of our attack methodology are provided in the Supplementary Material. Under this calibrated privacy risk, the diagnostic performance of the DP method dropped to 61.41\% AUROC, while ROFI maintained its performance of 91.25\% AUROC. This result highlights the significant utility cost incurred by standard DP approaches for medical diagnostic tasks.}

	{We further conducted a human face recognition study to assess privacy protection from a human perspective. The study was designed as a forced-choice matching task involving 5 participants, with detailed methodology provided in the Supplementary Material. In an initial phase with 200 trials (supplementary Table 10), a Kruskal-Wallis H-test confirmed a statistically significant difference in protection effectiveness among the various methods (H=21.69, P=0.0006). However, we noted that these initial trials were insufficient to robustly discriminate the performance of ROFI from the two other strong anonymization methods, Digital Mask and G2Face.
		
	Consequently, to validate our findings with greater statistical power, we conducted an additional 1,000 trials focusing specifically on these three top-performing methods. The expanded results (supplementary Table 10) showed that ROFI achieved a recognition rate of only 1.1\% (14/1200), demonstrating a substantially stronger performance than both Digital Mask at 3.5\% (42/1200) and G2Face at 3.1\% (38/1200). A one-tailed Fisher's Exact Test confirmed that ROFI's superiority is highly statistically significant when compared to both Digital Mask (P = 0.000099) and G2Face (P = 0.000529). These results provide strong statistical evidence that ROFI offers a more effective defense against human recognition.}

	ROFI supports reversibility. With the authorized neural identity restorer software and a private key, the original image can be accurately reconstructed from the ROFI image. Compared to G2Face, the only reversible method examined, our approach achieved superior reversed ID similarity (97.19\%) and image similarity (98.47\%) over G2Face's 74.72\% and 93.48\%, respectively (Figure~\ref{fig:face_recognition}d).
	The reversed image enabled reliable traceability for medical audits, as well as facilitating the precise retrieval of personalized medical records, thereby enhancing longitudinal examinations.
	
	Finally, we assessed the efficacy of TED hormone treatment by comparing the current image with a retrieved historical image (Figure~\ref{fig:face_recognition}e).
	In the ``Single'' scenario, where no historical information was utilized, the $\kappa$ value obtained was 0.2758 (95\% CI: -0.0860-0.6378). Using the general reversible privacy protection technique G2Face, $\kappa$ was only marginally improved to 0.3913 (95\% CI: 0.0956-0.6869), due to the suboptimal reconstruction quality of G2Face. Discrepancies in the reconstructed images, including alterations in skin color and increased blurriness (Figure~\ref{fig:face_recognition}f), impeded accurate retrieval of historical data, resulting in a less reliable comparison with the current image.
	In contrast, our method, ROFI, yielded the highest $\kappa$ value of 0.8888 (95\% CI: 0.7393-1.0384), attributable to its superior reconstruction performance.

	\section{Discussion}

	In this article, we introduced, developed, and validated a facial privacy protection technique ROFI for the ophthalmic department.
	ROFI learned from real patient data to prioritize the preservation of ophthalmic sign features that are crucial for the clinical diagnosis, while simultaneously obscuring identifying biometrics of patients.
	By harnessing the potent capabilities of deep learning in autonomous disease sign detection and identification protection, fueled with a large-scale real ophthalmic patient dataset, ROFI demonstrated superior performance in both clinical diagnosis and privacy protection.

	ROFI proved to be highly reliable for clinical diagnosis. A comprehensive clinical diagnosis study showed the remarkable consistency (Cohen's kappa $\kappa>0.81$) achieved by ophthalmologists when utilizing ROFI-protected images, compared to diagnostic results obtained from the original images.
	In comparison to previous privacy protection methods that does not consider any ophthalmic features or consider only hand-crafted features, our approach significantly outperformed them, particularly for diseases that rely on discriminating local subtle cues, such as Basal Cell Carcinoma, Conjunctival Melanoma, and Corneal Leukoma.

	{ROFI was shown to effectively obfuscate biometric features, thereby providing a high level of empirical privacy protection.} The results revealed that the identities of over 95\% of patients were successfully obfuscated, preventing recognition by facial recognition systems.
	The privacy protection rates were comparable to previous facial anonymous techniques, albeit with the latter achieving this by sacrificing some ophthalmic features that are necessary for diagnosing eye diseases. Since our method automatically identified and preserved only the disease-related ocular regions, it avoided introducing additional privacy risks, even with the retention of more complete ophthalmic features.
	
	{
		With different private keys, ROFI generates diverse outputs.
		we randomly sample 100 different keys to encrypt the same input image into distinct results. We then measure the standard deviation of the encrypted images in both the pixel space and the identity feature space.
		As shown in the supplementary Table 14, varying key leads to a substantial pixel-wise deviation ($\sigma_{\text{pixel}} = 36.52$), indicating that the generated images are visually distinct. Moreover, it results in a large mean pairwise cosine distance of 0.8845 between their identity features.
		We further visualize the patient face images encrypted by different private key. As shown in Supplementary Figure 10, the different noises leads to different encryption results.
		This probabilistic generation results ensure that ROFI does not merely anonymize faces, but actively obfuscates them within a broad distribution of possible outputs, significantly increasing the difficulty and cost of any potential privacy attack.
	}

	ROFI offers several advantages over the traditional eye cropping approach. 
	First, regarding privacy protection, conventional eye cropping may expose more identifiable cues, such as periorbital skin texture and iris details—leading to only a 70\% protection rate, which is significantly lower than the 95\% efficacy achieved by ROFI, as demonstrated in Figure~5 (C).
	
	In terms of iris recognition attacks, ROFI remains effective as it preserves disease-relevant information, such as overall iris shape and color, while obfuscating identity-sensitive iris details like pigment distribution. To evaluate iris privacy, we used the open-source iris recognition algorithm \texttt{OpenIris}~\cite{github2024openiris}. The original images (or those processed by eye cropping) had a 74.45\% recognition success rate, highlighting the vulnerability of standard facial images to iris-based identification. In contrast, ROFI-protected images achieved only a 0.87\% success rate, indicating that identity-related iris features are effectively obscured by our method.

	Second, regarding disease diagnosis, eye cropping removes surrounding facial features entirely, while ROFI preserves key facial structures. This enables the observation of clinically relevant facial manifestations, thereby supporting more comprehensive medical assessments. For instance, in patients with Bell’s palsy~\cite{baugh2013clinical}, our ROFI framework enables accurate detection of facial landmarks, which can be used to assess facial asymmetry—a key clinical indicator of the condition. Such diagnostically valuable features would be completely lost when using eye-only cropping strategies.
	
	Even in cases of ocular diseases, certain conditions such as squamous cell carcinoma (SCC) can present with symptoms extending beyond the eye region, potentially involving the forehead or cheeks~\cite{bernstein1996many}.
	Certain specific full-face features, such as ``hypothyroid facies''~\cite{uday2014hashimoto}, are also helpful for diagnosing thyroid eye disease (TED).
	Our ROFI framework retains these extraocular signs, whereas eye cropping discards them entirely, potentially delaying timely medical follow-up.
	
	Finally, we mention that ROFI is not contradictory to eye cropping. Rather, ROFI is fully compatible with subsequent eye cropping, offering ophthalmologists the flexibility to focus on the ocular region or adopt this combination as a safer and more informative strategy.

	ROFI had been demonstrated to be effective when integrated into medical AI models. These models are adopted in the preliminary screening of diseases, significantly conserving physician resources while broadening the scope of early disease detection.
	Moreover, given that most AI models are deployed on remote cloud platforms, owing to their high computational demands and reliance on GPU clusters, there is an imperative need to protect the privacy of patient information contained within transmitted images. 
	Our findings indicated that AI diagnostic outcomes derived from ROFI-protected images exhibit substantial consistency ($\kappa > 0.81$) with those from the original images. {These findings indicate that ROFI effectively anonymizing patient identity with minimal impact on the medical AI's decision-making process, as evidenced by the high agreement rates.}

	ROFI was reversible in terms of both the identification and appearance of patient facial images.
	{For long-term patient care, such as monitoring chronic conditions like Thyroid Eye Disease, clinicians must compare current images to a historical baseline. ROFI’s key-based reversal restores this crucial longitudinal link, which is lost in irreversible methods.  Additionally, reversibility is essential for medical and legal auditing. It provides a secure ``digital fingerprint'' that ensures traceability and accountability, allowing auditors to verify that a diagnosis corresponds to the correct patient record, thus maintaining the integrity of clinical workflows.}
	Traceability was critical for maintaining accurate medical records and auditing the clinical treatment procedure, aligning with the GCP standard~\cite{world2005handbook,kern2006medical,grimes2005good}.
	Despite its reversibility, our method remained safe due to the confidentiality of both the protection and restoration software, which are distributed to partners only after signing a confidentiality agreement. Furthermore, even if the protection and decoder soft-wares are exposed to attackers, without the private key, the attacks can still not obtain the original image.
	
	{The privacy key-based encryption scheme of the ROFI model is safe enough.
		First, the key is the 512-dimensional float vector with 32-bit precision, providing $2^{512\times 32}  =2^{16384}$ possible combinations, which is computationally infeasible to brute-force.
		To empirically validate this, we conducted experiments where each encrypted sample from SNPH validation set was subjected to 1,000 brute-force attempts, showing a 0\% success rate, confirming the robustness against collision attack.
		Further, our approach can be combined with timestamp-based password authentication protocols, such as kerberos protocol~\cite{pirzada2004kerberos}, to defense replay attacks.
		
		We also attempted to obtain a universal key capable of decrypting any encrypted image through gradient-based optimization, in an adversarial manner. Specifically, given the encrypted images generated by ROFI, we input them into the decryption module of ROFI. We froze the parameters of the ROFI model and optimized only the key vector, with the objective of minimizing the difference between the decrypted image and the corresponding original image.
		With the optimized key, the protection rate of the decrypted data still remained quite high at 96.96\%.}

	ROFI is robust to adversial attack.
 We used the Basic Iterative Method (BIM) algorithm~\cite{kurakin2018adversarial} to generate adversarial noise, which was then added to the ROFI-protected images to perform the identity spoofing attack. The optimizing goal of the adversarial attack was to make the decrypted images resemble those of a spoofing identity.
	The iterative step is set to 15. Notably, even with a perturbation noise norm $\epsilon$ of 0.05, the attack success rate remains below 5\%.
	To further enhance robustness, we implemented a defense strategy by adding random noise to the input during inference~\cite{cohen2019certified}. After applying this defense mechanism, the attack success rate dropped to zero across all tested $\epsilon$ values (Supplementary Table 11), demonstrating the reliability of our approach. These results confirm the robustness of our method, whether or not the defense strategy is applied.
	
	{The implications of ROFI extend broadly across digital healthcare. It can accelerate the adoption of telemedicine by allowing patients to securely transmit clinically useful images without fear of identification, democratizing access to specialist care. For research, ROFI can help break down data silos by enabling the creation of large, multi-center, privacy-compliant datasets, which is critical for studying rare diseases and training robust AI models. Furthermore, by demonstrating that diagnostic models can be effectively trained on privacy-protected images, ROFI paves the way for developing trustworthy and ethical AI systems that can be safely deployed in cloud environments.}

	Despite the aforementioned advantages and contributions of ROFI to the privacy-protected clinical diagnosis field, it inevitably has some limitations.
	{First}, although ROFI was evaluated in three cohorts,
	the data was from Asian cohorts. In the future, we will validate the clinical outcomes on individuals from other racial cohorts.
	{Second}, ROFI is evaluated on facial images, which could be used to initially assess the external eye diseases such as ptosis. Future work will explore privacy exposure and protection issues associated with fundus photographs, which could be adopted in the diagnosis of other eye diseases such as the diabetic retinopathy.
	{{Third}, real-world deployment of ROFI presents ethical and technological challenges that must be addressed. Ethically, the most significant hurdle is key management and governance. Robust policies are needed to define who can authorize image reversal and under what circumstances, supported by a secure Key Management System with strict audit trails. The informed consent process must also be transparent, clearly explaining to patients how their data is protected and the reversible nature of the process. Technologically, challenges include seamless integration with existing hospital IT systems like PACS and EMR, and the need for sufficient computational resources (e.g., GPU clusters) to process image data at scale.}
	Fourth, in future, we will also evaluate the effectiveness of the proposed ROFI technique on recently emerging large multi-modal medical models~\cite{team2024gemma,aibench,zhang2025lmmsurvey,chen2025can,ji2025medomni,li2025information,li2025image}.
	
	In summary, we have substantiated the effectiveness of the proposed ROFI technique across various applications, including clinical diagnosis, privacy protection, compatibility to medical AI models, and longitudinal diagnosis support.
	ROFI exemplifies a critical step towards harmonizing the dual goals of advancing medical diagnostics with remote medicine and safeguarding patient privacy. As we move into an era where digital medicine plays an increasingly pivotal role, technologies like ROFI will be indispensable in building a future where healthcare is both effective and ethically sound.

	\section{Methods}
	
	The model development was approved by the Institutional Review Boards of Shanghai Ninth People's Hospital (Approval No.SH9H-2022-T380-1), Eye Center of Xiangya Hospital of Central South University (Approval No.202407131), and Renmin Hospital of Wuhan University (Approval No.WDRY2024-K238).
	All patients agreed to participate in the prospective study at the ROFI Program either by themselves or via their legal guidance.
	For the publication of identifiable images, the written informed consent was obtained from the parents or legal guardians.
	This study complies with the latest version of the Declaration of Helsinki~\cite{bierer2024declaration}.
	
	\subsection{Model Architecture of ROFI}

	{The ROFI framework is designed to achieve the dual goals of privacy and accuracy through a multi-stage process. First, an Ophthalmic Sign Detector, trained via weakly supervised learning, identifies and extracts clinically relevant sign features from the ocular regions. In parallel, a Transformer-based Neural Identity Translator alters the global facial features to obscure the patient's identity, guided by a private key. The framework then intelligently integrates these two outputs by replacing the eye-region features in the anonymized feature map with the original, sign-preserving features. A refinement network (DA-Former) ensures a seamless transition between these regions before a final decoder, which reconstructs a high-quality, privacy-protected image..}

	The whole architecture of the ROFI model is outlined in Figure~\ref{fig:model_overview}a, where all main components are implemented with neural networks and fully learnable.

	In the privacy protection procedure, ROFI employs the Ophthalmic Sign Detector and the Neural Identity Translator (Figure~\ref{fig:model_overview}b), to simultaneously preserve the ophthalmic features and protect the patient privacy.
	The Ophthalmic Sign Detector identifies disease-related features within eyes in a weakly-supervised manner, which will be detailed in Section~\ref{sec:oph_sign_det}. The Transformer-based Neural Identity Translator network alters the identifying facial appearances such as face shape.
	Subsequently, the output features from the above two components are integrated, i.e., the ophthalmic features are used to substitute the transformed facial features within the detected ophthalmic sign regions.
	The integrated feature map is then enhanced through a quality enhancement network DA-Former (Figure~\ref{fig:model_overview}c), amending the boundary defects caused by the feature substitution.
	The enhanced feature is then input into a decoding network Dec-Net (Figure~\ref{fig:model_overview}d), producing the privacy-protected facial images.
	During the privacy recovery procedure, we employ a Transformer-based Privacy Recovery network to revert the anonymized face to its original form, using the same private key as in the protection procedure.
	
	The Neural Identity Translator network (Figure~\ref{fig:model_overview}b) aims to alter the identity information within the facial image. The process begins with a feature extractor that transforms input image into the deep feature map, which consists of a series of convolution layers. The feature map is then flattened into a one-dimensional feature. {The private key is a 512-dimensional vector, drawn from the Gaussian distribution with unit variance, and is pre-pended before the flattened feature.} Then, the combined features are passed through six Transformer blocks~\cite{dosovitskiy2020image}, which leverage self-attention mechanism~\cite{vaswani2017attention} to transform the bio-identifying information within feature, effectively protecting the privacy. Finally, the output is unflattened back into the feature map.
	{
		We conduct ablation studies on the number of transformer blocks within the Neural Identity Transformer.
		As shown in Supplementary Table 12, the ID protection rate increases with the number of Transformer blocks, from 93.27\% using two blocks to 96.54\% with six blocks. Further increasing the depth to eight or ten blocks brings only marginal improvement (96.61\%), while significantly increasing computational cost and model size. Therefore, we adopt the six-block configuration as it achieves a good trade-off between performance and efficiency.
	}
	
	{
		We further visualize the facial features before/after the neural identity translation procedure.
		As shown in Figure~\ref{fig:vis_feat}, the original feature map highlights discriminative identity features such as the mouth and nose contours. After protection, the facial features are largely reduced or eliminated, while the eye features remain well-preserved and prominently highlighted.
	
}
	
	The architecture of the Neural Identity Restorer is similar to that of the Neural Identity Translator (Figure~\ref{fig:model_overview}b), with one difference: the input to the Neural Identity Restorer includes a learnable token that signals the network to operate in ``Restoring mode''.
	The learnable token is fixed across all patients.
	When the same private key used in the privacy protection procedure is applied, the original facial information could be restored. Conversely, if an incorrect private key is used, such as a random key by an attacker, the network generates the facial information of a virtual face, thereby maintaining patient privacy.
	This is achieved by introducing the learning objectives in Section~\ref{sec:framework_objective}.
	The wrong virtual face is determined for the same original face.
	This prevents attackers from exploiting output variability to infer patterns of the original image.
	
	The DA-Former module (Figure~\ref{fig:model_overview}c) aims to correct any artifacts caused by the stitching of the eye and facial features. The process begins with flattening the stitched feature map into a one-dimensional feature. The flattened feature is then passed through two Transformer blocks. After the transformations, the output is unflattened back into the feature map.
	The refinement network is guided by the GAN loss, which incorporates both local-wise and global-wise terms to guide the refinement process.
	The local-wise GAN loss guides the refined features to achieve a smooth transition between the eye and facial regions, while the global-wise loss enables that the refined information results in a realistic and coherent face image.
	
	The Dec-Net module (Figure~\ref{fig:model_overview}d) aims to reconstruct high-quality facial images from the feature map enhanced by DA-Former.
	It consists of a series of residue blocks~\cite{he2016deep} and upsample layers to gradually decode the high-resolution image from the low-resolution feature.
	{Concretely, the Dec-Net architecture sequentially comprises four residual blocks, an up-sampling layer, two residual blocks, another up-sampling layer, one residual block, and a final up-sampling layer. Each residual block consists of two 3$\times$3 convolutions with an intermediate ReLU activation function and skip connection. Batch normalization is not adopted to accurately preserve low-level features.}

	\subsection{Ophthalmic Sign Detector}
	\label{sec:oph_sign_det}
	The Ophthalmic Sign Detector is designed to detect ophthalmic sign features that are necessary for ophthalmic diagnosis (Figure~\ref{fig:sign_weak_learn}). For images from the SNPH cohort's developing set, physicians annotate each image's health state. A neural network $\phi$, which consists of an ophthalmic feature extractor followed by an ophthalmic sign scorer, scores each region within each image. The score of the region with the highest confidence score serves as the image-level disease confidence, supervised by an image-level binary label $y$. Specifically, $y=0$ indicates a healthy state, while $y=1$ denotes disease presence. This approach is termed the \textbf{``region-score-max''} learning strategy.
	
	Formally, given an input eye region image $X \in \mathbb{R}^{H \times W}$, $\phi$ assigns a sign score to each region of $X_{eye}$. The maximum abnormal score is used as the overall result for the entire image:
	\begin{equation}
		\begin{split}
			S &= \phi(X) \in \mathbb{R}^{\frac{H}{8} \times \frac{W}{8}},\\
			p& = \max(S), \\
			\mathcal{L}_{classify} &= - y \cdot \log(p) + (1 - y) \cdot \log(1 - p),
		\end{split}
	\end{equation}
	where $S$ represents the region-wise sign scores, and $p$ is the highest-confidence region score. As for the network architecture of $\phi$, ophthalmic feature extractor consists of the first three stages (Conv1, Res1, and Res2) of the ResNet50 architecture, while ophthalmic sign scorer is implemented as a linear layer followed by a Sigmoid function~\cite{narayan1997generalized}. 
	Despite training on image-level annotations, the detector effectively identifies classification-relevant regions in a weakly supervised manner, aligning with previous studies~\cite{hsu2019weakly,lu2021data}.
	Regions with a sign score exceeding 0.5 are designated as ophthalmic sign regions. Then, the features of these regions are selected as the ophthalmic feature.
	
	{The sign map threshold value 0.5 is determined by the empirical investigation.
		As shown in Table~\ref{tab:cutoff_comparison}, higher cutoff values (e.g., 0.7 or 0.9) result in more aggressive suppression of facial features, leading to better de-identification performance (ID Protection Rate = 96.96\%). However, such thresholds also risk removing diagnostically relevant regions, as evidenced by the noticeable drop in classification AUROC ( dropped to only 86.19\%).
		Conversely, lower cutoff values preserve more image content, which helps maintain high diagnostic accuracy (high AUROC), but leave identifiable features partially intact, reducing de-identification effectiveness.
		For example,  the ID protection rate is only 92.04\%, when the cutoff value is set to 0.1.
	}

	{Moreover, we visualize the sign maps to provide a more intuitive understanding of the ophthalmic features learned by the model.
		As shown in Figure~\ref{fig:vis_mask}, the disease-related ophthalmic signs are successfully captured.
		Although some false-positive regions are detected, they lack specific identity information and therefore do not compromise the overall de-identification performance.}
		
	\begin{table}[!thbp]
		\normalsize
		\centering
		\setlength{\tabcolsep}{5mm}
		
		\begin{tabular}{|c|c|c|c|c|c|}
			\hline
			{Threshold Value} & 0.1 & 0.3 & 0.5 & 0.7 & 0.9 \\ \hline
			{ID Protection Rate (\%)} & 92.04 & 93.93 & 96.54 & 96.96 & 96.96 \\ \hline
			{Classification AUROC (\%)} & 91.34 & 91.34 & 91.25 & 88.26 & 86.19 \\ \hline
		\end{tabular}
		\vspace{3mm}
		\caption{{Comparison of different sign map threshold values on the SNPH set. ID protection rate and classification AUROC is evaluated using the AdaCos face recognition and ViT classification networks.}}
		\label{tab:cutoff_comparison}
	\end{table}

	\subsection{Learning Objectives of the Framework}
	\label{sec:framework_objective}
	We denote the original face image and the protected face image as $X$ and $Y$.
	The recovered face images by correct password and wrong password are denoted as $\hat{X}$ and $X_{wrong}$, respectively.
	The learning objective includes five parts.
	
	(1) The cosine similarity between the identification of the protected image $Y$ and that of the original image $X$ should be as small as possible,
	\begin{equation}
		\mathcal{L}_{removeID} = \cos( \operatorname{Face-Net}(X),\operatorname{Face-Net}(Y) ),
	\end{equation}
	where $\operatorname{Face-Net}$ denotes a pre-trained face identification network Sphereface~\cite{liu2017sphereface}.
	
	(2) The identification of the correctly recovered image $\hat{X}$ and the original image $X$ should be as similar as possible, namely, the cosine similarity should be large,
	\begin{equation}
		\mathcal{L}_{recoverID} = - \cos( \operatorname{Face-Net}(\hat{X}),\operatorname{Face-Net}(X) ).
	\end{equation}
	
	(3) The appearance of the correctly recovered image $\hat{X}$ and the original image $X$ should be as similar as possible,
	\begin{equation}
		\mathcal{L}_{recoverApp} =L1( \hat{X}-X)*0.05 + LPIPS( \hat{X}, X), 
	\end{equation}
	where L1 denotes the $\ell1$ norm function, LPIPS denotes the Learned Perceptual Image Patch Similarity metric~\cite{zhang2018unreasonable}.
	
	(4) The identification of the wrongly recovered image ${X}_{wrong}$ and the original image $X$ should be as dissimilar as possible.
	\begin{equation}
		\mathcal{L}_{wrongID} = \cos( \operatorname{Face-Net}({X}_{wrong}),\operatorname{Face-Net}(X) ).
	\end{equation}
	
	(5) All images should be photo-realistic.
	\begin{equation}
		\mathcal{L}_{photo} = \mathcal{L}_{GAN}(X) + \mathcal{L}_{GAN}(Y)  +\mathcal{L}_{GAN}(\hat{X}) + \mathcal{L}_{GAN}(X_{wrong}), 
	\end{equation}
	where $ \mathcal{L}_{GAN}$ denotes the Generative Adversarial Networks (GAN)~\cite{goodfellow2020generative} loss, which forces the images to be visually natural.
	{Concretely, the GAN discriminator architecture follows a PatchGAN design~\cite{isola2017image} with five 4×4 convolutional layers (stride=2) for progressive downsampling. Each convolution is followed by spectral normalization and LeakyReLU activation functions. The GAN loss comprises two components: (1) a local naturalness constraint applied to patch-level features, and (2) a global naturalness constraint implemented through average pooling across all spatial positions.
	}
	{To stabilize GAN training, we introduce the following techniques: spectral normalization~\cite{miyato2018spectral} layers to control discriminator Lipschitz constants, and Two-Timescale Update Rule (TTUR)~\cite{heusel2017gans} to balance generator and discriminator learning rates.}
	
	{The GAN framework plays a crucial role in ensuring that the prepended noise key is actively utilized rather than ignored. The Generator is incentivized to produce a wide variety of diverse and realistic outputs to fool the Discriminator. Given our large-scale training dataset, the noise key serves as an effective perturbation signal, which the Generator leverages as a source of randomization to create distinct facial identities for the same input image. As demonstrated in our quantitative analysis (Supplementary Table 14 and Supplementary Figure 10), different noise keys lead to high variance in both pixel space and identity feature space, confirming that the model actively uses the noise to generate diverse, protected images.}
	
	Finally, the while objective is given by,
	\begin{equation}
		\mathcal{L}_{ROFI} = \mathcal{L}_{removeID} +\mathcal{L}_{recoverID} +\mathcal{L}_{recoverApp} + \mathcal{L}_{wrongID} +\mathcal{L}_{photo}.
	\end{equation}

	\subsection{Training Strategy}
	During the training of the ROFI model, we utilize the AdamW optimizer~\cite{kingma2014adam} with the following hyper-parameters: $\beta_1$ (0.5), $\beta_2$ (0.999), a weight decay of 1e-8, and a learning rate of 1e-4. The batch size is set to 12, and the training iteration number is 1,000,000. The learning rate undergoes a decay of 10 times at the 900,000th iteration.
	We evaluated the model every 20,000 iterations on the model-selection set of SNPH, the model with the smallest loss value $\mathcal{L}_{ROFI}$ is adopted as the final model.
	As the data augmentation strategy, we employ the random horizontal flipping operation.
	The model is implemented with the PyTorch framework~\cite{paszke2019pytorch}.
	{During the training of the ophthalmic sign detector, the optimizer and hyper-parameters are same as that of ROFI. To prevent overfitting to dominant regions (e.g., sclera/iris) and allows for detecting of subtle signs like early-stage conjunctival melanoma (CM), we employ the following strategies: (1) Weight Decay~\cite{loshchilov2017decoupled}: This regularization technique discourages over-reliance on dominant features by penalizing large network weights, promoting balanced feature learning across both prominent and subtle regions. (2) Random Crop Augmentation~\cite{takahashi2019data}: By training the model on randomly cropped sub-regions of the eye image, we force it to focus on localized discriminative cues rather than global dominant patterns. This enables the model sensitive to fine-grained abnormalities that may appear in non-dominant areas.}

	\subsection{Implementation of the AI Diagnostic Models}
	
	The SNHP validation set is partitioned into three subsets: the training set (60\%) is utilized for training the AI models, the validation set (10\%) is employed for hyperparameter tuning, and the test set (30\%) serves as the benchmark for evaluating model performance.
	Throughout the training phase, all images undergo a uniform resizing process, bringing them to a standardized dimension of $256 \times 256$ using cubic interpolation. Subsequently, these resized images undergo augmentation employing prevalent strategies, including random cropping (ranging from 30\% to 100\% of the whole image), resizing to $224 \times 224$, random horizontal flipping, and image normalization.
	The ECXHCSU validation set is partitioned into two subsets, the training set (60\%) and the test set (40\%).
	ECXHCSU has no hyperparameter tuning subset, since its hyperparameters are reused from SNHP.
	
	The weight decay values for ResNet50 and ViT networks are configured at 1e-4 and 1e-2, respectively. The learning rate is set to 1e-4 for both ResNet50 and ViT. Mini-batch size is set to 256 for both ResNet50 and ViT. The training iteration number is set to 3000. Both ResNet50 and ViT networks are initialized from the models pretrained on ImageNet dataset~\cite{deng2009imagenet}.
	We optimize the diagnostic models with the AdamW optimizer~\cite{loshchilov2017decoupled}. During model evaluation, all images are resized to $256 \times 256$ and subsequently cropped to the central $224 \times 224$ region.
	{The strategy was adopted from the official data-processing strategies of the above network architectures~\cite{he2016deep}~\cite{dosovitskiy2020image}}

	\section*{Data Availability}
	The data supporting the findings of this study are divided into two categories: shared data and restricted data.
	Shared data are available in the manuscript, references, and supplementary information.
	The source data of all figures and tables will be released to publicly-accessible project website upon publication.
	Restricted data such as the patient photos, which pertain to individual participants, are subject to a license that permits their use solely for analysis. Consequently, these data cannot be publicly shared.
	Access to patient photos is available upon reasonable request and the submission of a formal data usage agreement. All requests are subject to review and approval by the relevant ethics and data governance committees. 
	
	\section*{Code Availability}
	The codes and the trained model are available at \url{https://github.com/tianyuan168326/PatientFaceAnonymizer-Pytorch}.
	Data was analysed with python 3.10, cudatoolkit 11.6.0, pytorch 2.1.2, numpy 1.24.3, opencv-python 4.8.1.78, pandas 2.0.3, scipy 1.10.1, pillow 10.0.1, piq 0.8.0, and
	dlib 19.24.2.

	\section*{Acknowledgments}
	The work is supported by National Natural Science Fund of China (62501337), National Natural Science Fund of China (82388101), National Natural Science Fund of China (72293585), National Natural Science Fund of China (72293580), and Shanghai Eye Disease Research Center (2022ZZ01003).
	
	\section*{Author Contributions}
	All authors have read and approved the final version of the manuscript. The contributions of each author are as follows.
	X. Fan, G. Zhai, X. Song: Designed the study framework, proposed the hypothesis, and oversaw the integration of multi-omics data. Secured funding and provided critical infrastructure for data acquisition.
	Y. Tian, M. Zhou: Conducted the experiments, performed statistical analysis, and drafted the manuscript.
	H. Zhou, Y. Chen, G. Zhai: Developed analytical pipelines for clinical data, performed image annotation, and ensured cross-center harmonization.
	F. Li, L. Qi, X. Xu, Y. Yu: Completed data curation, investigation, and contributed to clinical data collection.
	S. Xu, C. Lei, J. Tan, L. Wu, H. Chen, X. Liu, W. Lu, L. Li: Coordinated patient enrollment and multi-center clinical data collection.
	Y. Jiang, R. Zhang, S. Wang: Cleaned the data, converted data into model-training format, designed the network architecture of the ROFI model, trained the model, and validated its performance.
	
	\section*{Competing Interests}
	The authors have no competing interests as defined by Nature Portfolio.

	\bibliographystyle{naturemag}
	\bibliography{references}
	
			\begin{figure*}[!thbp]
		\centering
		\vspace{-4mm}
		\setlength{\tabcolsep}{1mm}
		\begin{minipage}[t]{\linewidth}
			\begin{minipage}[t]{0.99\linewidth}
				\begin{minipage}[t]{\linewidth}
					
					\begin{minipage}[t]{0.34\linewidth}
						\centering
						\begin{tabular}{c}
							\multicolumn{1}{l}{\textsf{\small \textbf{a)}}} \\
							\includegraphics[width=\linewidth]{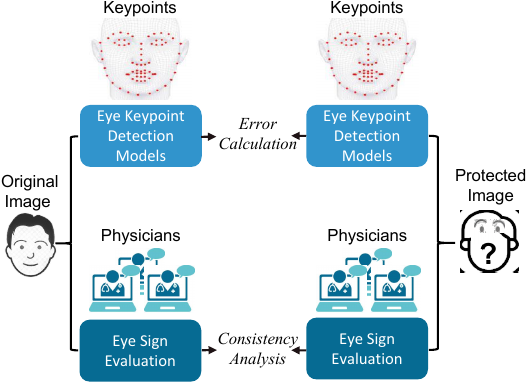}
						\end{tabular}
					\end{minipage}
					\hfill
					\begin{minipage}[t]{0.45\linewidth}
						\centering
						\begin{tabular}{cc}
							\multicolumn{2}{l}{\textsf{\small \textbf{b)}}} \\
							\vspace{-5mm}
							\includegraphics[width=0.48\linewidth]{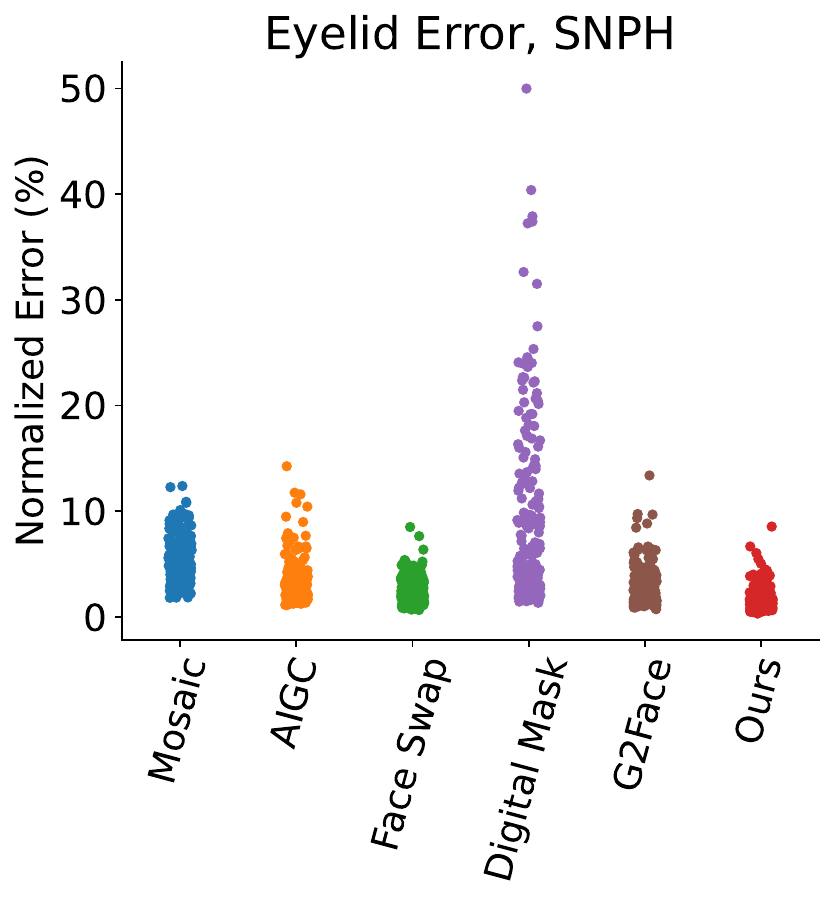} &
							\includegraphics[width=0.48\linewidth]{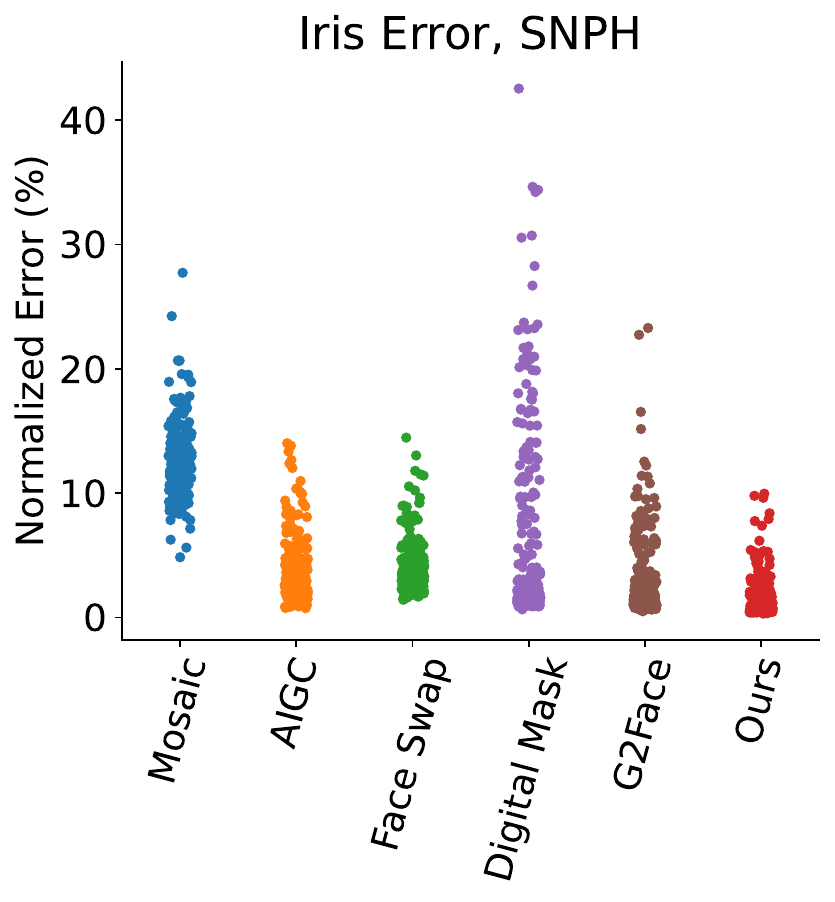}
						\end{tabular}
					\end{minipage}%
					\hfill
					\begin{minipage}[c]{0.16\linewidth}
						\centering
						\begin{tabular}{c}
							\multicolumn{1}{l}{\textsf{\small \textbf{c)}}} \\
							\includegraphics[width=1\linewidth]{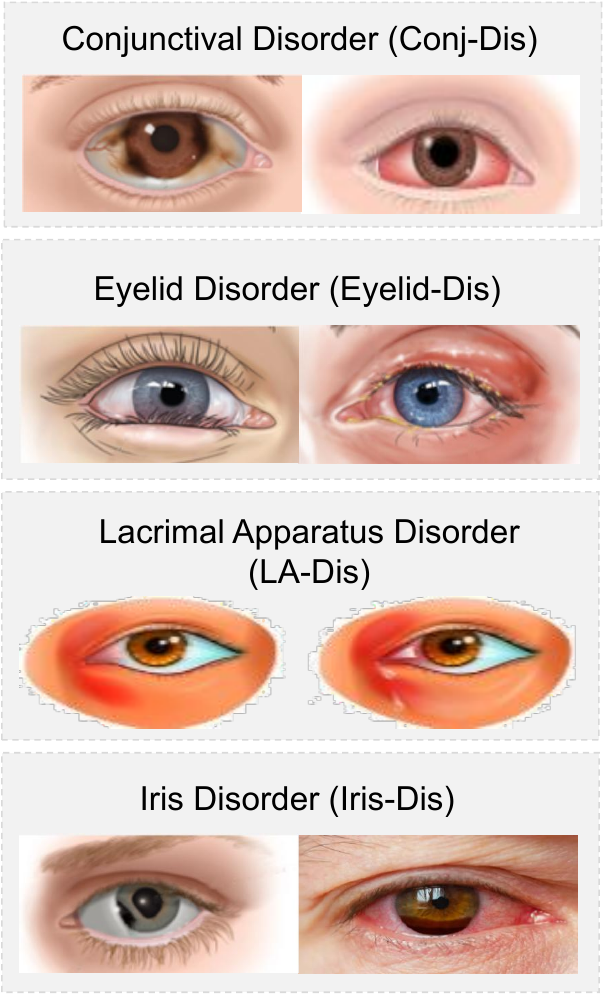} 
						\end{tabular}
					\end{minipage}
				\end{minipage}
				\begin{minipage}[t]{\linewidth}
					\begin{minipage}[t]{1\linewidth}
						\begin{minipage}[t]{\linewidth}
							\vspace{0pt} %
							\centering
							\begin{tabular}{ccc}
								\multicolumn{3}{l}{\textsf{\small \textbf{d)}}} \\
								\includegraphics[width=0.35\linewidth]{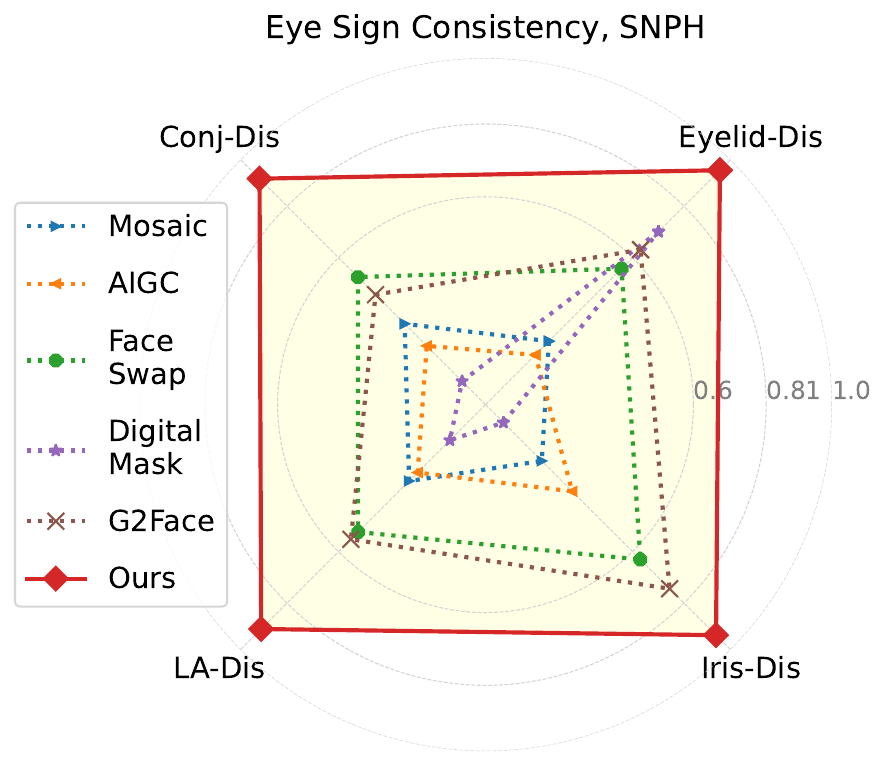} &
								\includegraphics[width=0.30\linewidth]{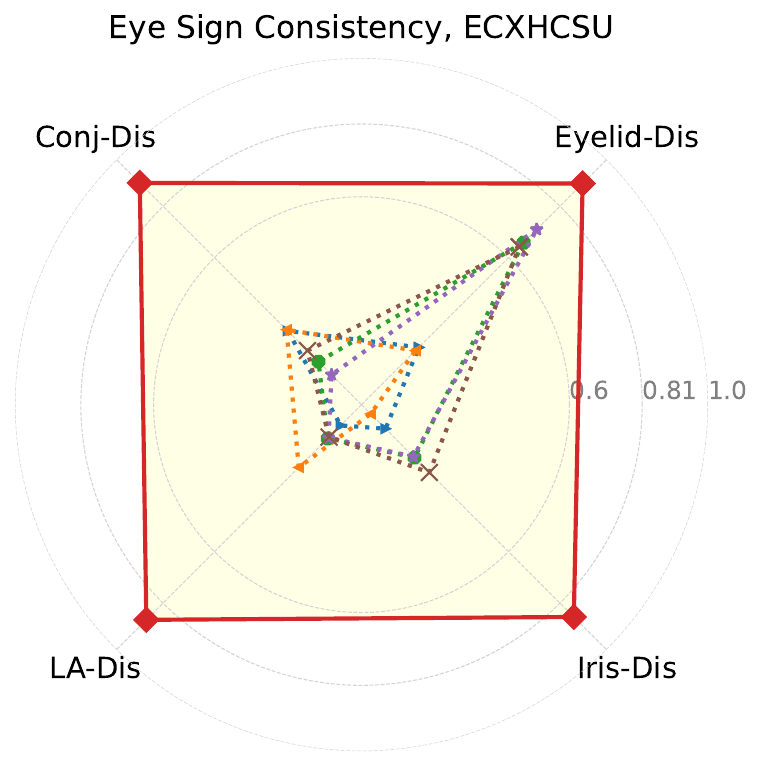} &
								\includegraphics[width=0.30\linewidth]{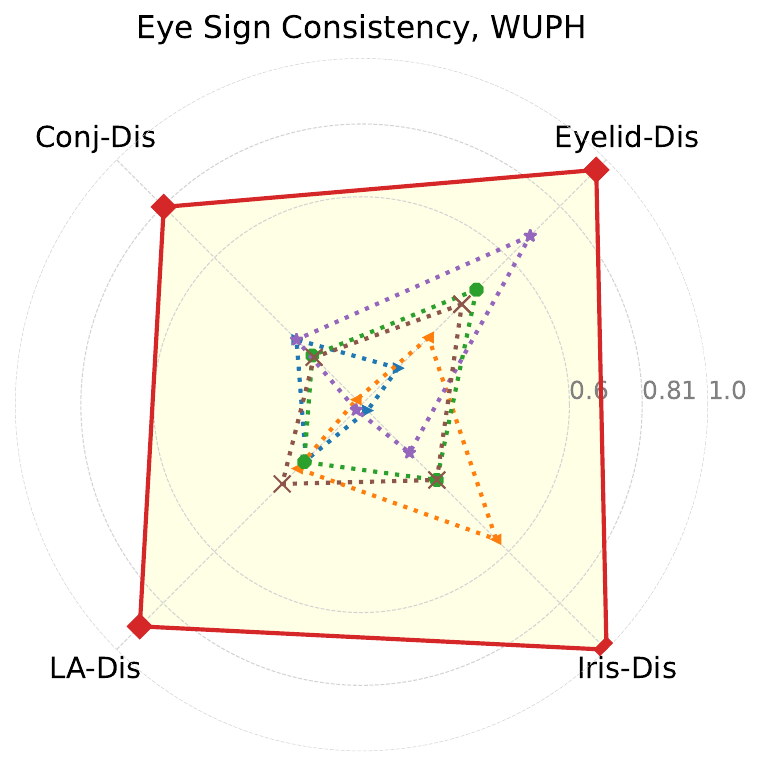} 
							\end{tabular}
						\end{minipage}
						\begin{minipage}[t]{\linewidth}
							\begin{minipage}[t]{0.40\linewidth}
								\sffamily
								\vspace{0pt} % Ensure top alignment
								\centering
								\scalebox{0.7}{ % 这里0.8是缩放比例，可以根据需要调整
									\begin{tabular}{|c|c|c|c|c|}
										\hline
										P-Value &Conj-Dis & Eyelid-Dis & LA-Dis & Iris-Dis \\
										\hline
										Ours-Mosaic&$< 0.001 $ &$ < 0.001 $&$ < 0.001 $ &$ < 0.001 $ \\
										Ours-AIGC&$< 0.001 $ &$ < 0.001 $&$ < 0.001 $ &$ < 0.001 $ \\
										Ours-Face Swap&$< 0.001 $ &$0.0350 $&$ < 0.001 $ &$ < 0.001 $ \\
										Ours-Digital Mask&$< 0.001 $ &$ < 0.001  $&$ < 0.001 $ &$ < 0.001 $ \\
										Ours-G2Face&$< 0.001 $ &$0.0197 $&$ < 0.001 $ &$ < 0.001 $ \\
										\hline
									\end{tabular}
								}
							\end{minipage}
							\hfill
							\begin{minipage}[t]{0.29\linewidth}
								\sffamily
								\vspace{0pt}
								\centering
								\scalebox{0.7}{ % 这里0.8是缩放比例，可以根据需要调整
									\begin{tabular}{|c|c|c|c|}
										\hline
										Conj-Dis & Eyelid-Dis & LA-Dis & Iris-Dis \\
										\hline
										$< 0.001 $ &$< 0.001 $&$ < 0.001 $ &$ < 0.001 $ \\
										$< 0.001 $ &$< 0.001 $&$ < 0.001 $ &$ < 0.001 $ \\
										$< 0.001 $ &$< 0.001 $&$ < 0.001 $ &$ < 0.001 $ \\
										$< 0.001 $ &$< 0.001 $&$ < 0.001 $ &$ < 0.001 $ \\
										$< 0.001 $ &$< 0.001 $&$ < 0.001 $ &$ < 0.001 $ \\
										\hline
									\end{tabular}
								}
							\end{minipage}
							\hfill
							\begin{minipage}[t]{0.29\linewidth}
								\sffamily 
								\vspace{0pt} 
								\centering
								\scalebox{0.7}{ % 这里0.8是缩放比例，可以根据需要调整
									\begin{tabular}{|c|c|c|c|}
										\hline
										Conj-Dis & Eyelid-Dis & LA-Dis & Iris-Dis \\
										\hline
										$< 0.001 $ &$< 0.001 $&$ < 0.001 $ &$ < 0.001 $ \\
										$< 0.001 $ &$< 0.001 $&$ < 0.001 $ &$ < 0.001 $ \\
										$< 0.001 $ &$< 0.001 $&$ < 0.001 $ &$ < 0.001 $ \\
										$< 0.001 $ &$< 0.001 $&$ < 0.001 $ &$ < 0.001 $ \\
										$< 0.001 $ &$< 0.001 $&$ < 0.001 $ &$ < 0.001 $ \\
										\hline
									\end{tabular}
								}
							\end{minipage}
						\end{minipage}
						\begin{minipage}[t]{\linewidth}
							%						\vspace{-5mm}
							\centering
							\begin{tabular}{c}
								\multicolumn{1}{l}{\textsf{\small \textbf{e)}}} \\
								\includegraphics[width=1\linewidth]{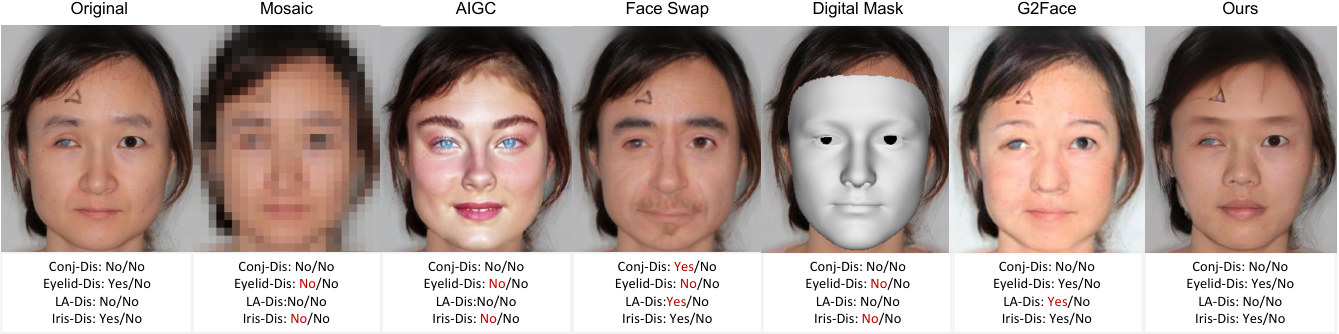} 
							\end{tabular}
						\end{minipage}
						
					\end{minipage}
				\end{minipage}
			\end{minipage}
		\end{minipage}
		\caption{
			\textbf{Eye Sign-Level Evaluation.}
			\textbf{a)} Workflow of the eye sign evaluation procedure.
			Eye keypoints were obtained from both original and protected images.
			The Mean Squared Errors (MSEs) of these keypoints were then calculated. Additionally, physicians annotated four common ophthalmic signs, namely, abnormal conjunctiva, abnormal eyelid, abnormal sclera, and abnormal iris, for both original and protected images. Then, the result consistency was assessed.
			\textbf{b)} Keypoint measurements on the SNPH validation set in terms of eyelid and iris error. Results on the ECXHCSU and WUPH validation sets are detailed in the Supplementary Figure 1.
			The mean and the standard deviation (SD) values are also provided in the Supplementary Table 1. 
			Per-disease results are provided in Supplementary Figure 3 and Supplementary Figure 4.
			\textbf{c)} Illustration of the eye signs evaluated in our study.
			\textbf{d)} Eye sign consistency (Cohen's $\kappa$) on the SNPH, ECXHCSU, and WUPH validation sets.
			$k \textgreater 0.81$ indicates remarkable consistency~\cite{dettori2020kappa}.
			$P$ values between our result and the results achieved by other approaches were calculated with the two-sided McNemar's test~\cite{pembury2020effective}.
			Detailed results and 95\% Confidence Intervals (CIs) are provided in Supplementary Table 3.
			Icons are from https://uxwing.com/ and https://www.biorender.com/.
			Created with BioRender.com.
			{	Conj-Dis:conjunctival disorder;
				LA-Dis: lacrimal apparatus disorder;
				Eyelid-Dis: eyelid disorder;
				Iris-Dis: iris disorder.}
			\textbf{e,} Qualitative comparison of the eye signs of the images processed by different privacy protection methods.
		}
		\label{fig:quan_results}
		%		\vspace{-1cm}
	\end{figure*}
	
		\begin{figure*}[!thbp]
		\centering
		\setlength{\tabcolsep}{1mm}
		\begin{minipage}{0.80\textwidth}
			
			\begin{minipage}{1\textwidth}
				\centering
				\begin{tabular}{c}
					\multicolumn{1}{l}{\textsf{\small \textbf{a)}} } \\
					\includegraphics[width=0.98 \textwidth]{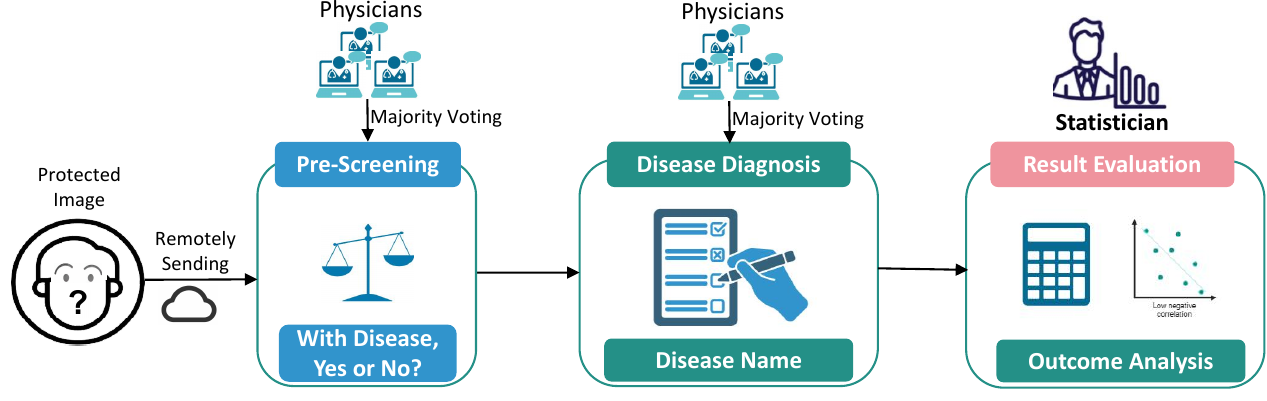}
				\end{tabular}
			\end{minipage}
			
			\begin{minipage}	{1\linewidth}
				\vspace{-3mm}
				\begin{tabular}{ccc}
					
					\multicolumn{3}{l}{\textsf{\small \textbf{b)}} } \\
					\includegraphics[width=0.32 \linewidth]{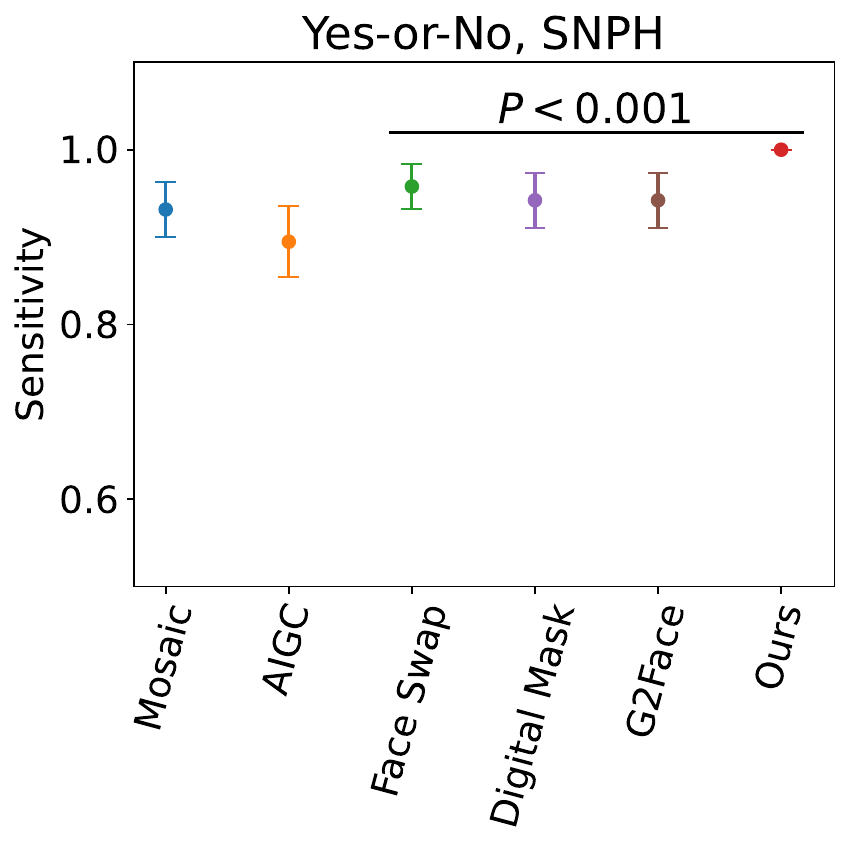}&
					\includegraphics[width=0.32 \linewidth]{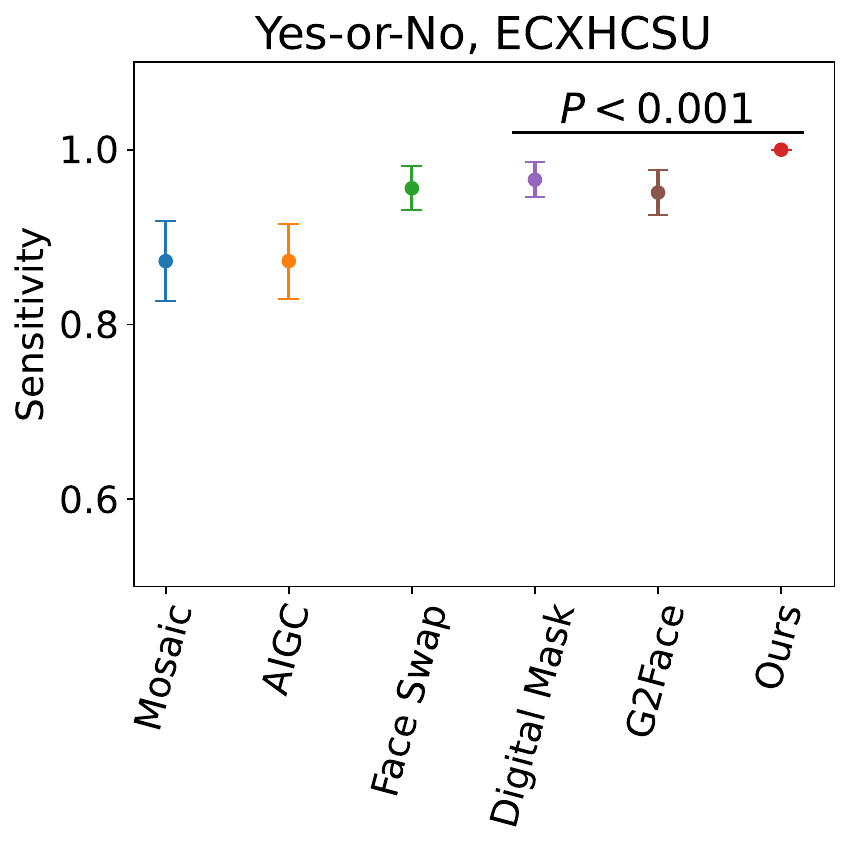}&
					\includegraphics[width=0.32 \linewidth]{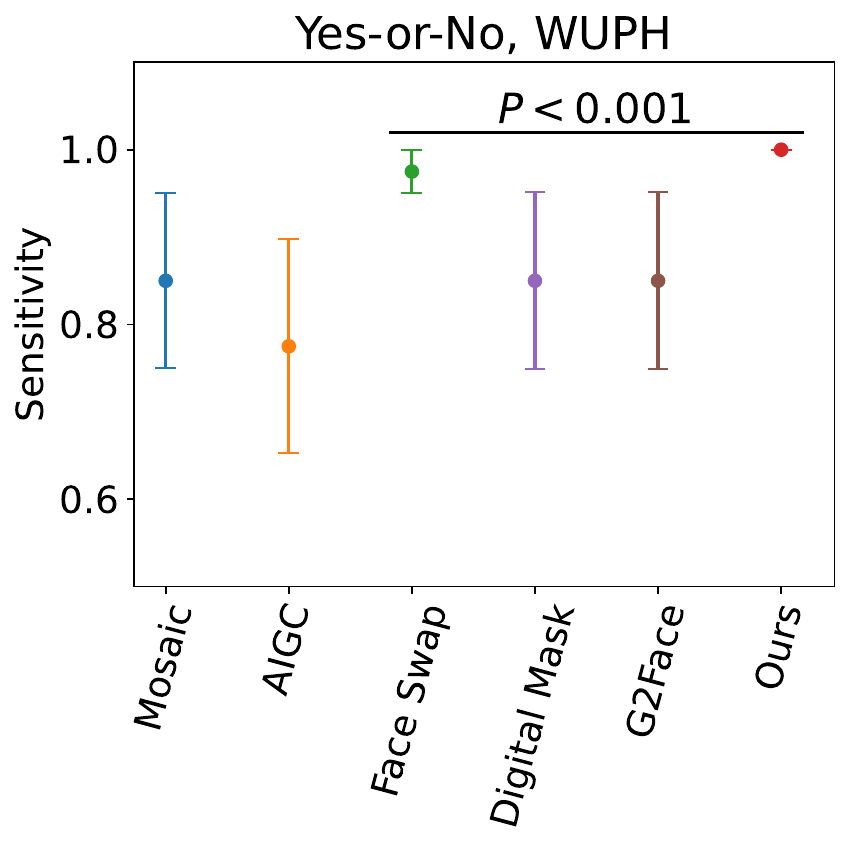}
					\\
				\end{tabular}
			\end{minipage}

			\begin{minipage}	{1\linewidth}
				\begin{tabular}{ccc}
					\multicolumn{3}{l}{\textsf{\small \textbf{c)}} } \\
					\includegraphics[width=0.39 \linewidth]{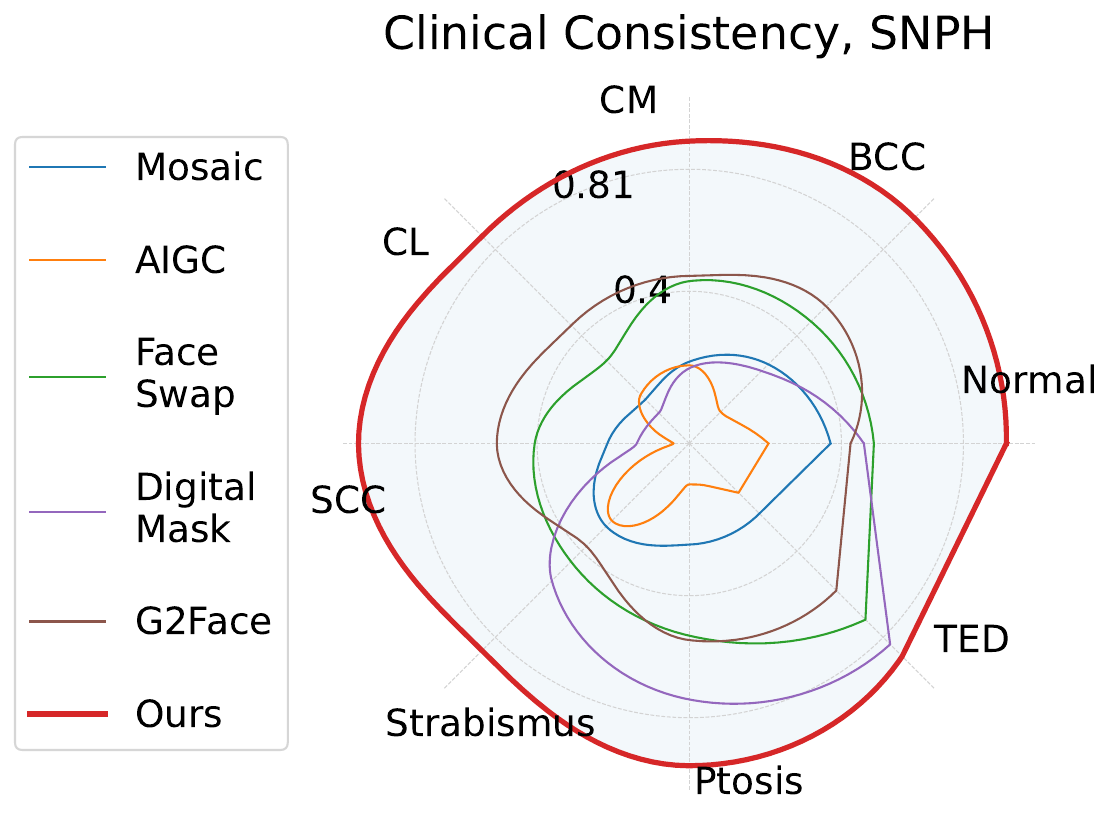}&
					\includegraphics[width=0.29 \linewidth]{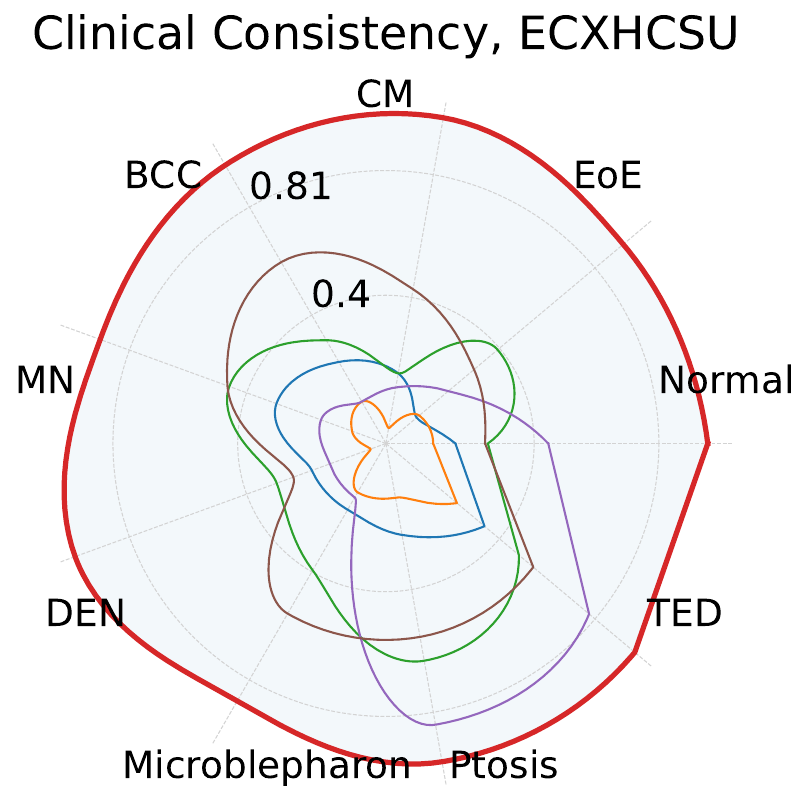}&
					\includegraphics[width=0.28 \linewidth]{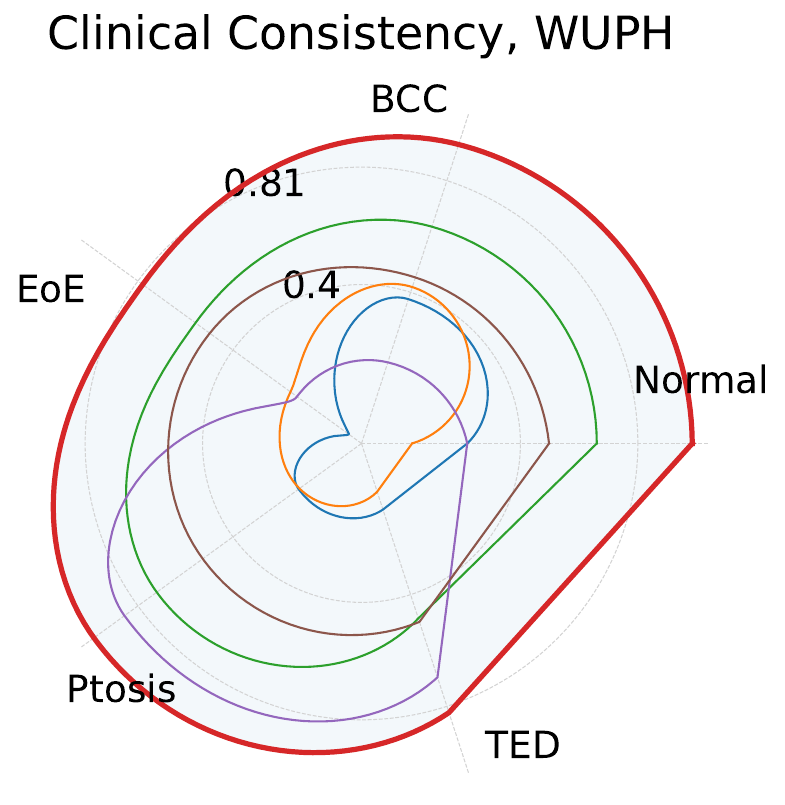}
					\\
					\makecell[c]{\small Ours $\kappa$=0.9376 \\ \small (95\% CI: 0.9033-0.9719)}&
					\makecell[c]{\small Ours $\kappa$=0.9621 \\ \small(95\% CI: 0.9346-0.9896)}&
					\makecell[c]{\small Ours $\kappa$=0.9433 \\ \small(95\% CI: 0.8674-1.0192)}
					\\
				\end{tabular}
			\end{minipage}
		\end{minipage}
		\begin{minipage}{0.19\textwidth}
			\renewcommand\arraystretch{3}
			\begin{tabular}{c}
				\multicolumn{1}{l}{\textsf{\small \textbf{d}} } \\
				\includegraphics[width=1 \linewidth]{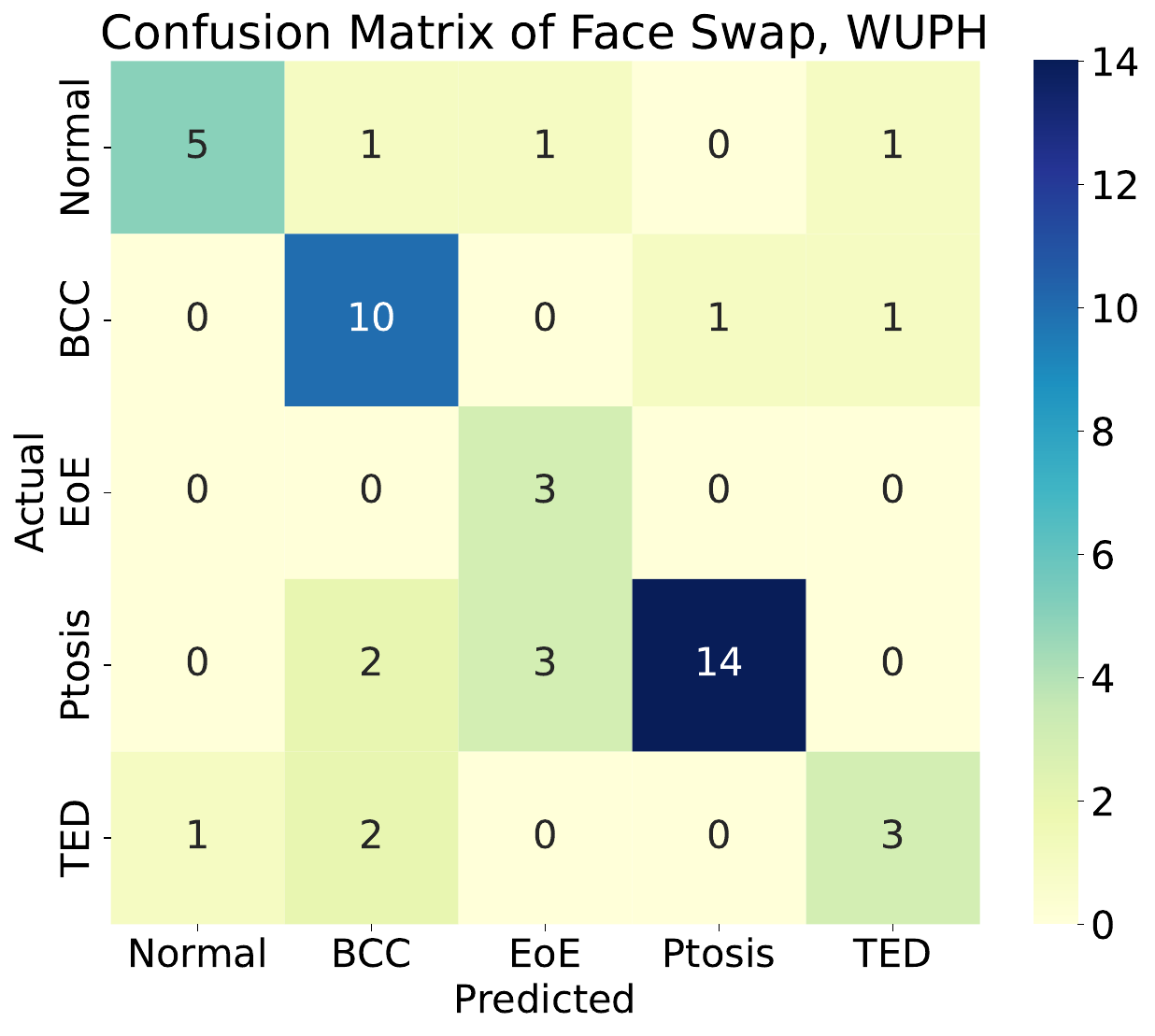}\\
				\includegraphics[width=1 \linewidth]{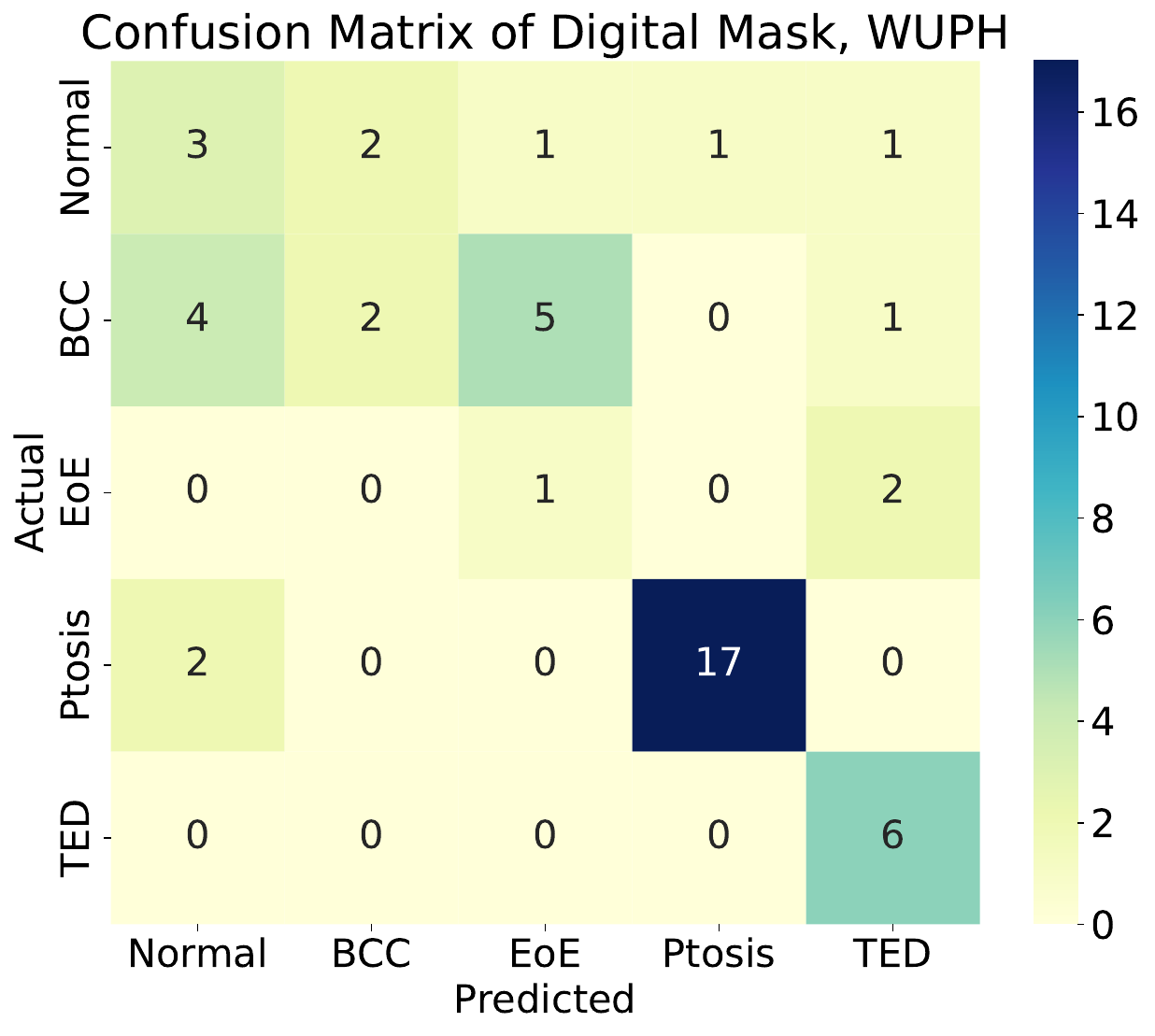}\\
				\includegraphics[width=1 \linewidth]{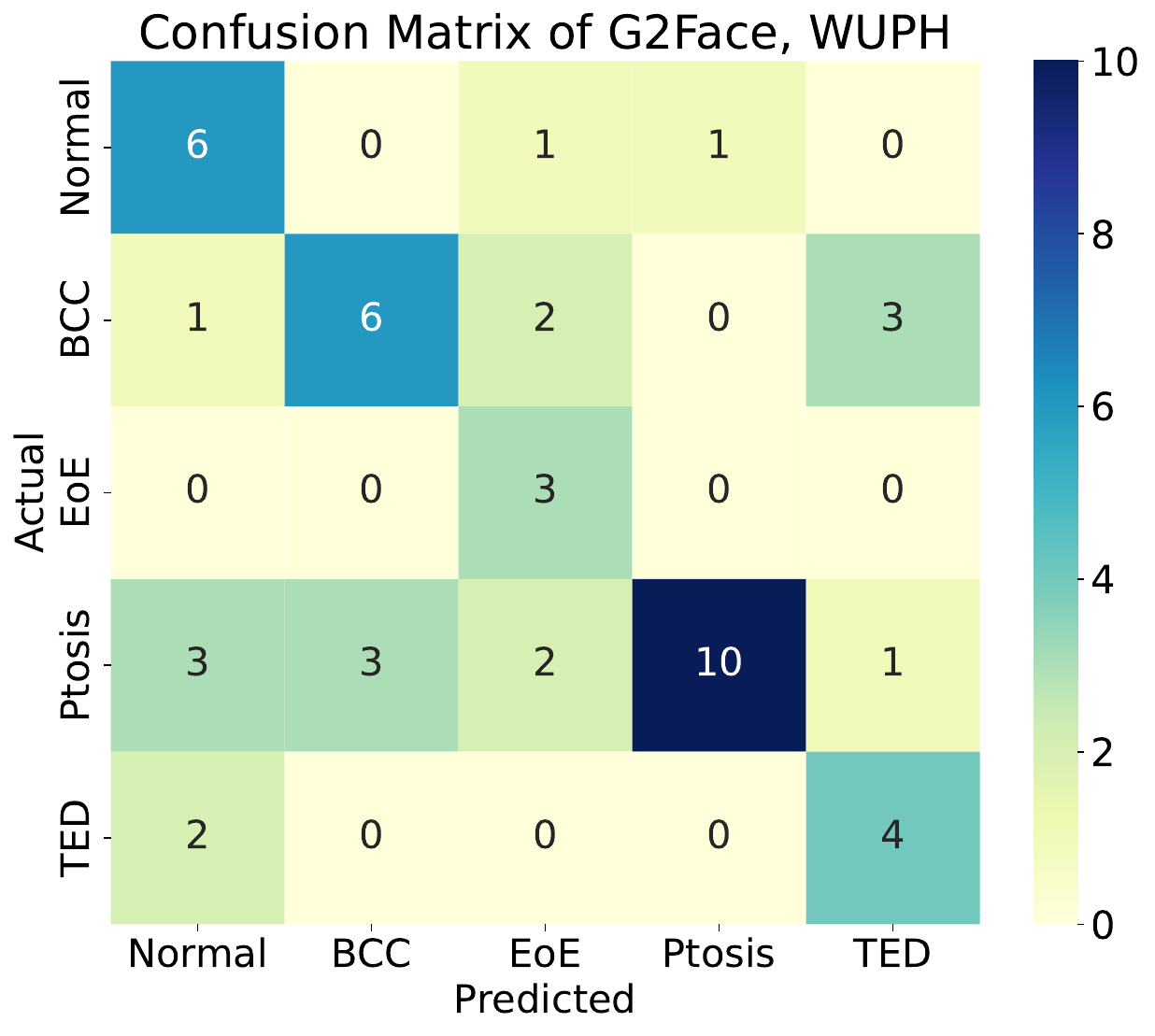}\\
				\includegraphics[width=1 \linewidth]{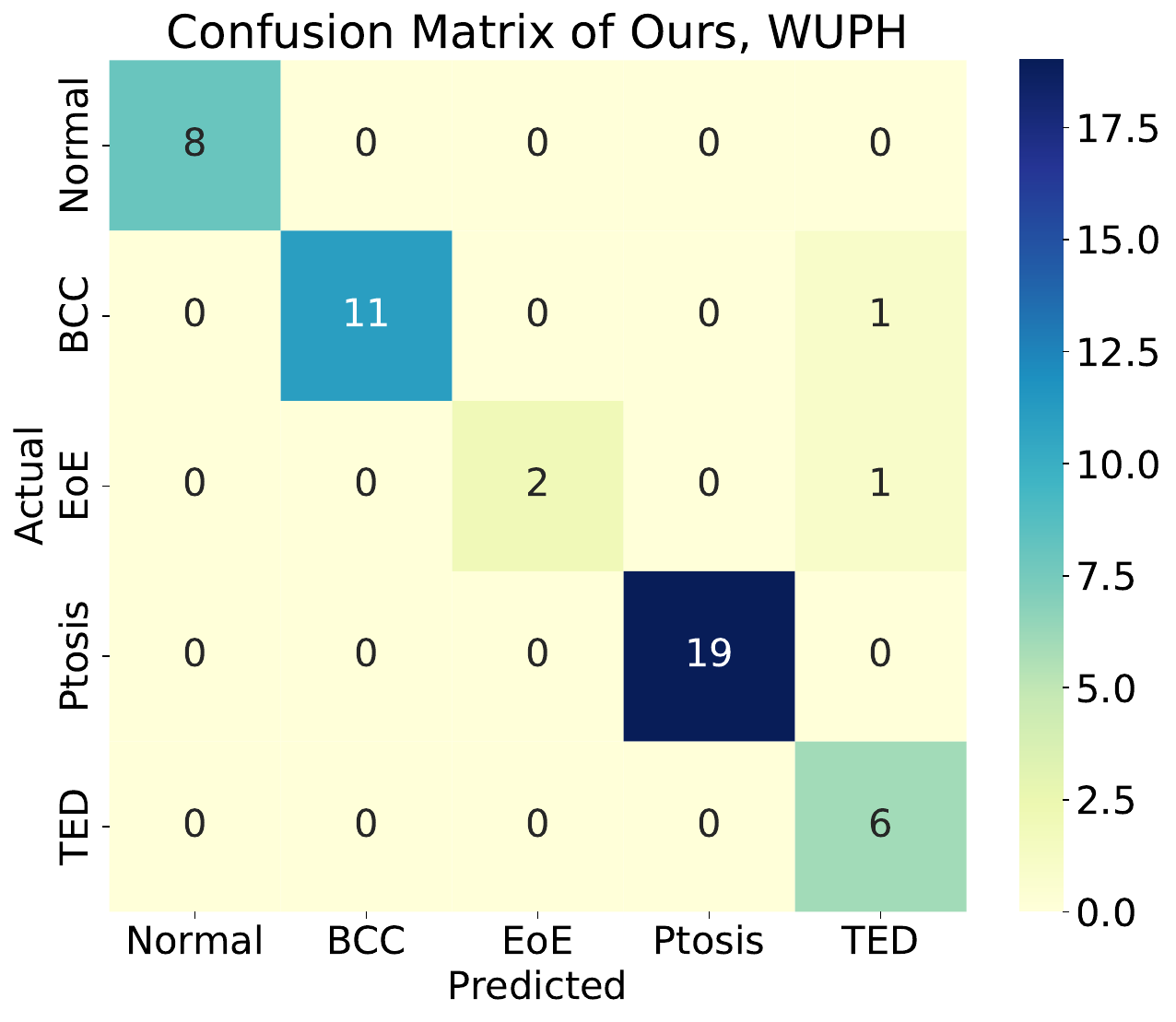}
				\\
			\end{tabular}
		\end{minipage}
		
		\caption{
			\textbf{Eye Disease-Level Evaluation.}
			\textbf{a)} 
			Workflow of the diagnostic evaluation procedure.
			Three ophthalmologists first judged whether each image is with disease or not (Yes-or-No). The final result was determined by majority voting.
			The same three physicians then diagnosed the specific disease by selecting from a predefined list. If more than two physicians provided the same result, that result was chosen. If all three physicians provided different results, the final result was determined through a consultation.
			The above procedure was repeated for all privacy protection methods and all validation sets.
			Finally, the results were analyzed by a statistician.
			\textbf{b)} Scatter plot indicating the sensitivity of the diagnosis results on images protected by different methods.
			Details are provided in Supplementary Table 4.
			\textbf{c)}
			Radar plots illustrating diagnostic consistency (Cohen's $\kappa$) for the specified ophthalmic diseases. Curves of varying colors represent different privacy protection methods. A value of $\kappa \geq 0.81$ indicates excellent diagnostic consistency~\cite{dettori2020kappa} between the original and privacy-protected images.  
			Detailed results and 95\% Confidence Intervals (CIs) of $\kappa$ are provided in Supplementary Table 5.
			The Matthews Correlation Coefficient (MCC)~\cite{chicco2020advantages} comparison of different methods are provided in Supplementary Table 6 and Supplementary Figure 4.
			We also report 
			{BCC: basal cell carcinoma;  
				SCC: squamous cell carcinoma;  
				CM: conjunctival melanoma;  
				CL: corneal leukoma;  
				TED: thyroid eye disease;  
				EoE: entropion or ectropion;  
				MN: melanocytic nevi;  
				DEN: divided eyelid nevus.}
			\textbf{d)} Confusion matrix comparison of different methods on WUPH.
			The confusion matrices on SNPH and ECXHCSU are provided in Supplementary Figure 5 and Supplementary Figure 6.
			Icons are from https://uxwing.com/ and https://www.biorender.com/.
			Created with BioRender.com.
		}
		\label{fig:trail_clinical}
	\end{figure*}

	\begin{figure*}[!thbp]
		\begin{minipage}[t]{0.95\linewidth}
			
			\begin{minipage}[t]{0.49\linewidth}
				\setlength{\tabcolsep}{0.1mm}
				\begin{tabular}{cc}
					\multicolumn{2}{l}{\textsf{\small \textbf{a)}} } \\
					%				\\\vspace{-5mm}
					\includegraphics[width=0.50 \linewidth]{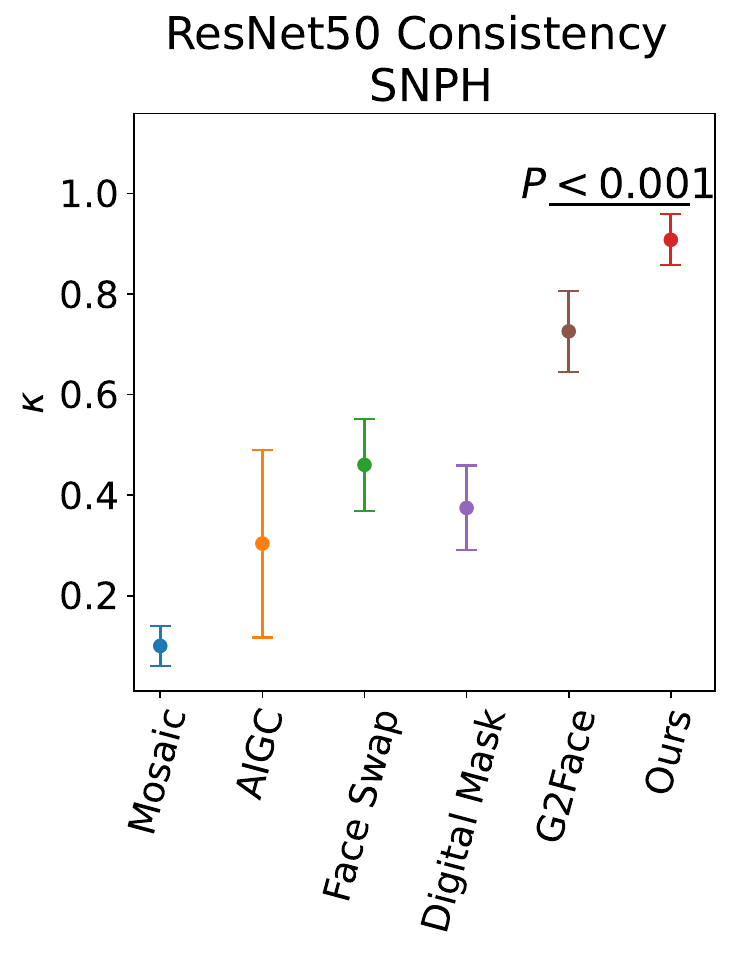}&
					\includegraphics[width=0.50 \linewidth]{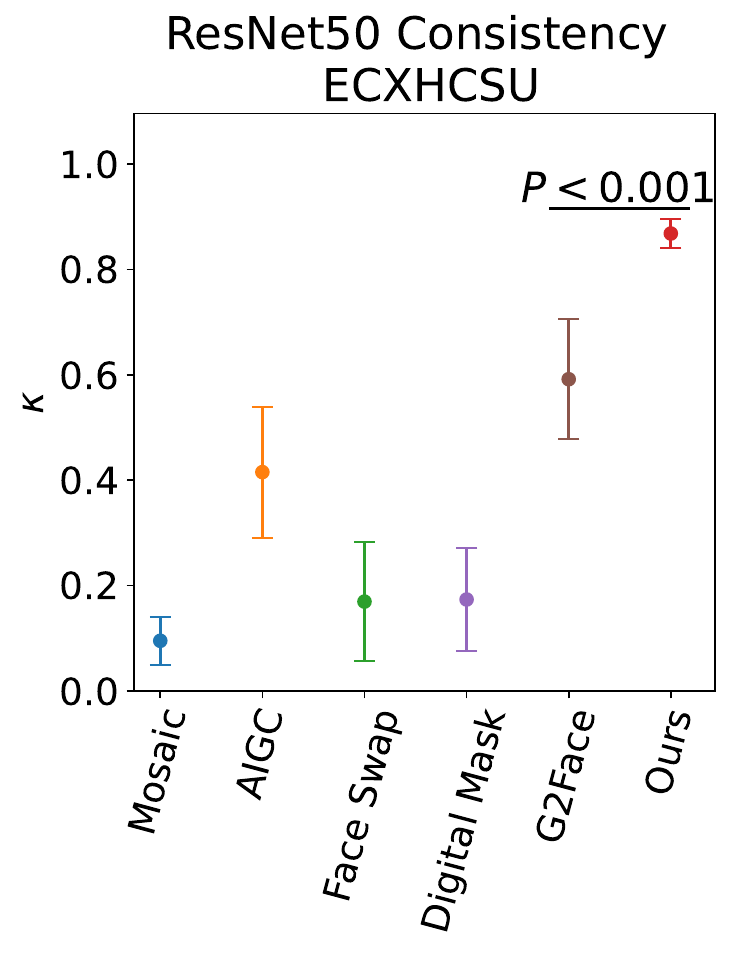}
				\end{tabular}
			\end{minipage}
			\hfill
			\begin{minipage}[t]{0.49\linewidth}
				\setlength{\tabcolsep}{0.1mm}
				\begin{tabular}{cc}
					\multicolumn{2}{l}{\textsf{\small \textbf{b)}} } \\
					\includegraphics[width=0.50 \linewidth]{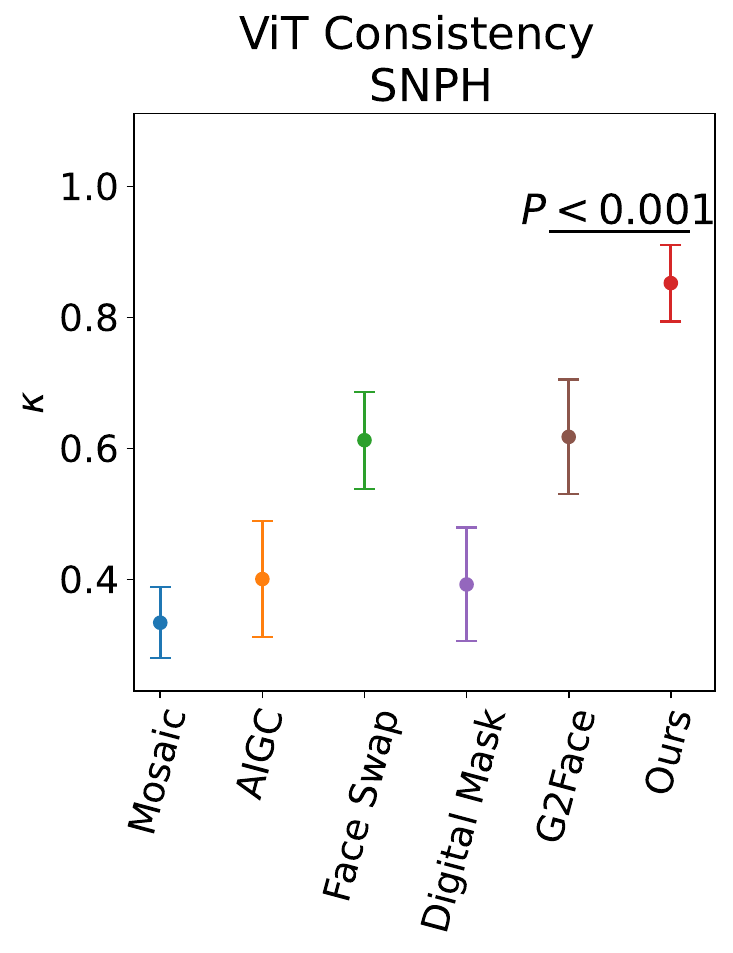}&
					\includegraphics[width=0.50 \linewidth]{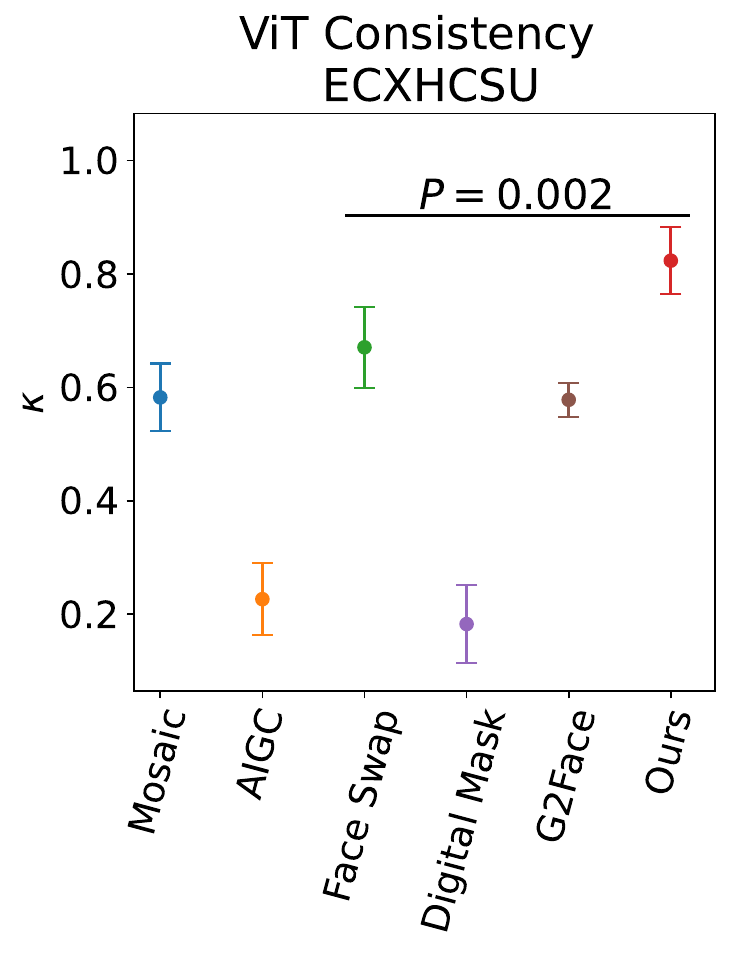}
				\end{tabular}
			\end{minipage}
			
			\begin{minipage}[t]{1\linewidth}
				\setlength{\tabcolsep}{0.1mm}
				\begin{tabular}{cccc}
					\multicolumn{4}{l}{\textsf{\small \textbf{c)}} } \\
					\includegraphics[width=0.25 \linewidth]{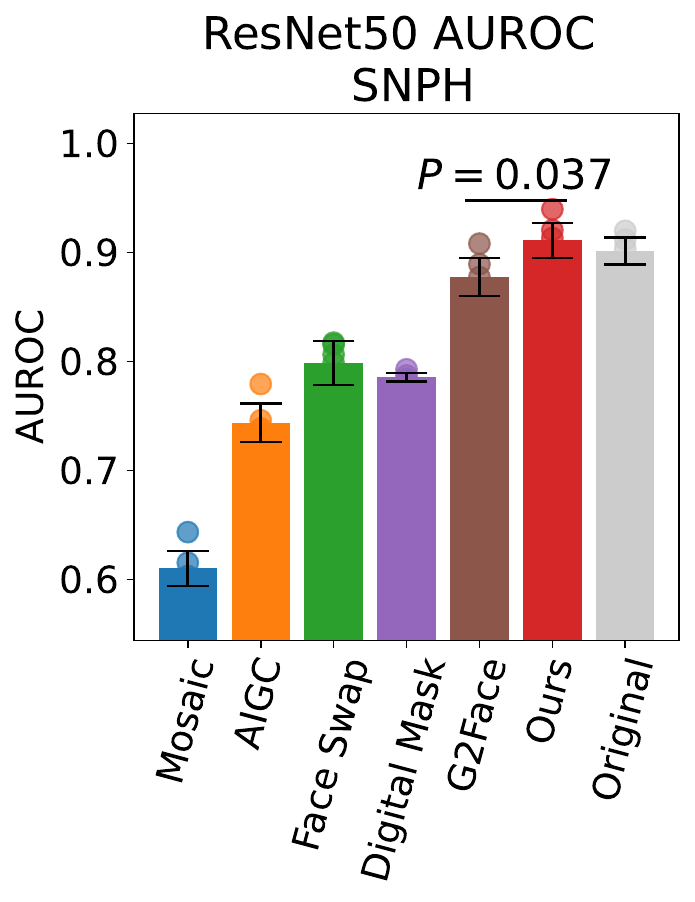}&
					\includegraphics[width=0.25 \linewidth]{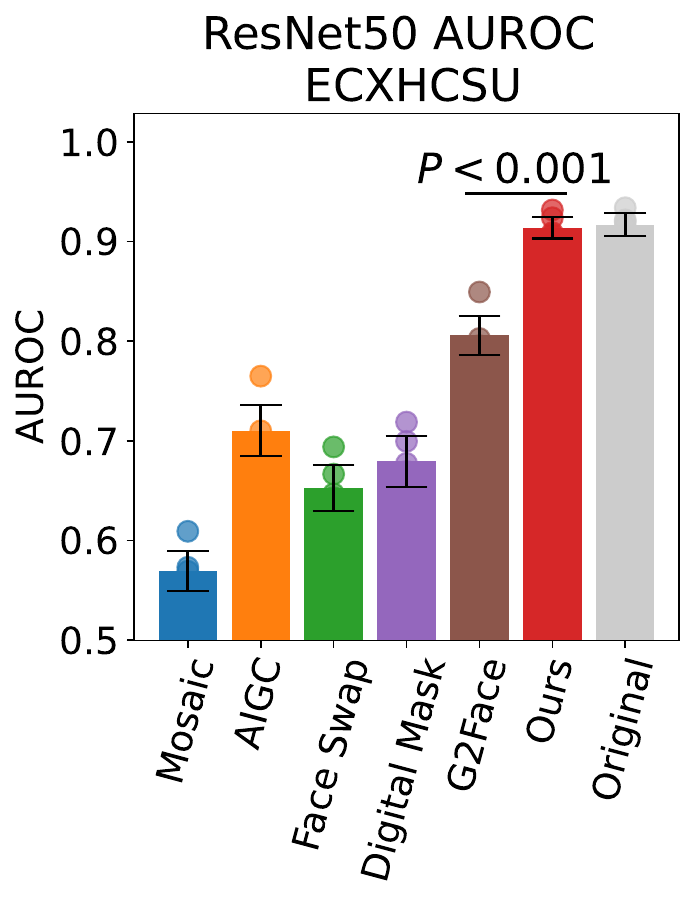}&
					\includegraphics[width=0.25 \linewidth]{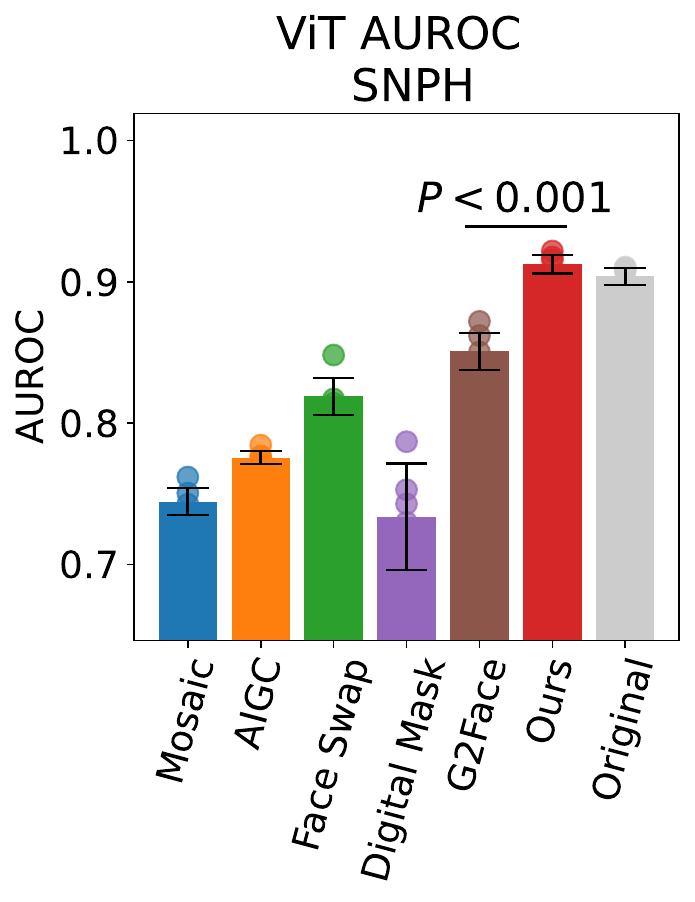}&
					\includegraphics[width=0.25 \linewidth]{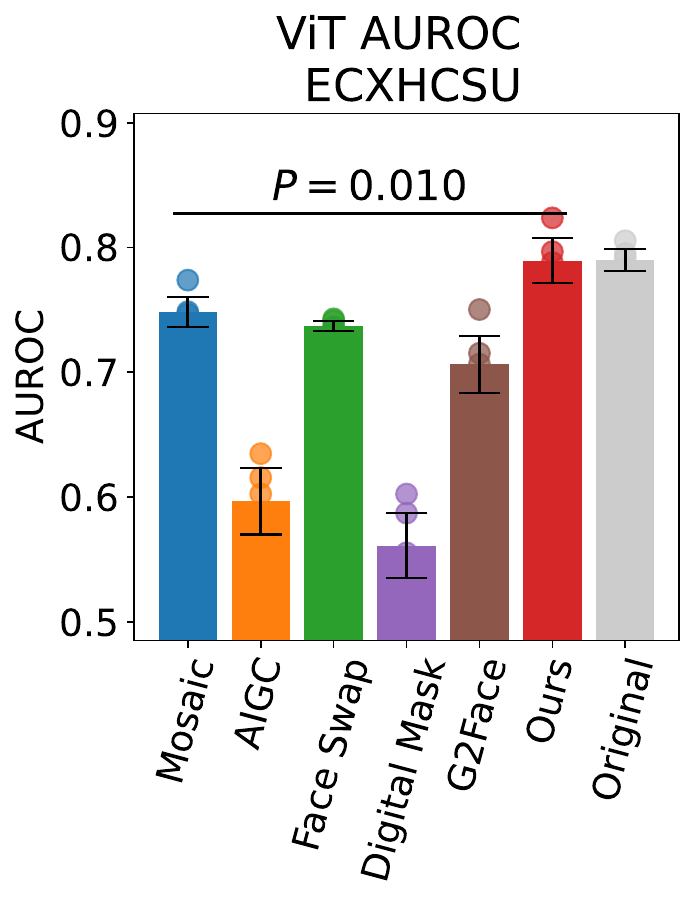}
				\end{tabular}
			\end{minipage}
			
			\begin{minipage}[t]{0.95\linewidth}
				\setlength{\tabcolsep}{0.5mm}
				\begin{tabular}{cccc}
					\multicolumn{4}{l}{\textsf{\small \textbf{d)}} } \\
					\includegraphics[width=0.255 \linewidth]{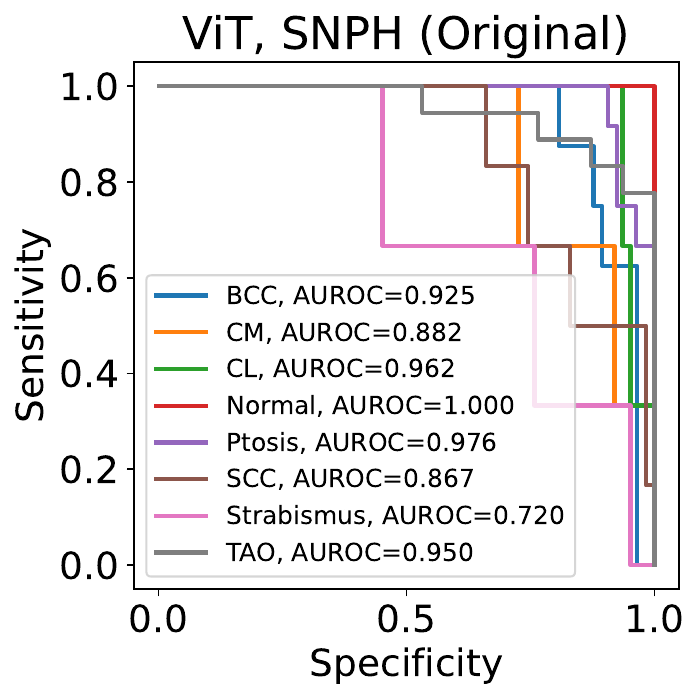}&
					\includegraphics[width=0.255 \linewidth]{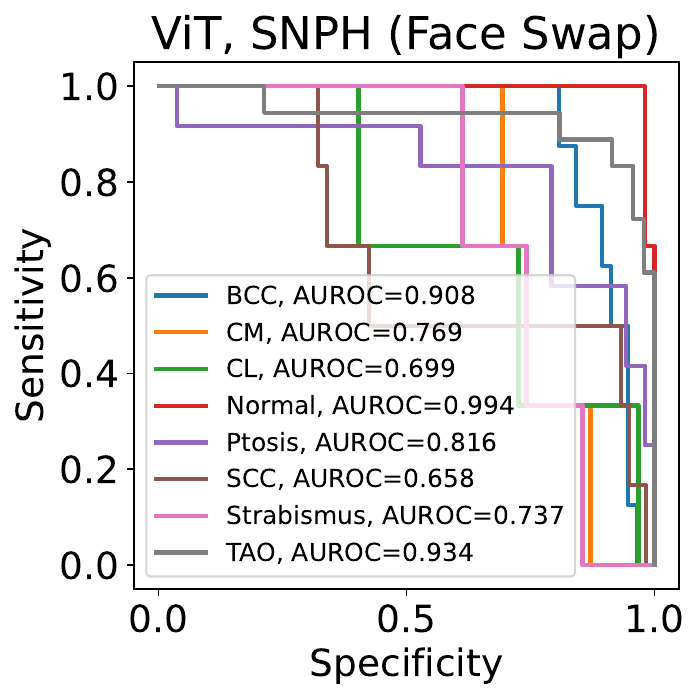}&
					\includegraphics[width=0.255 \linewidth]{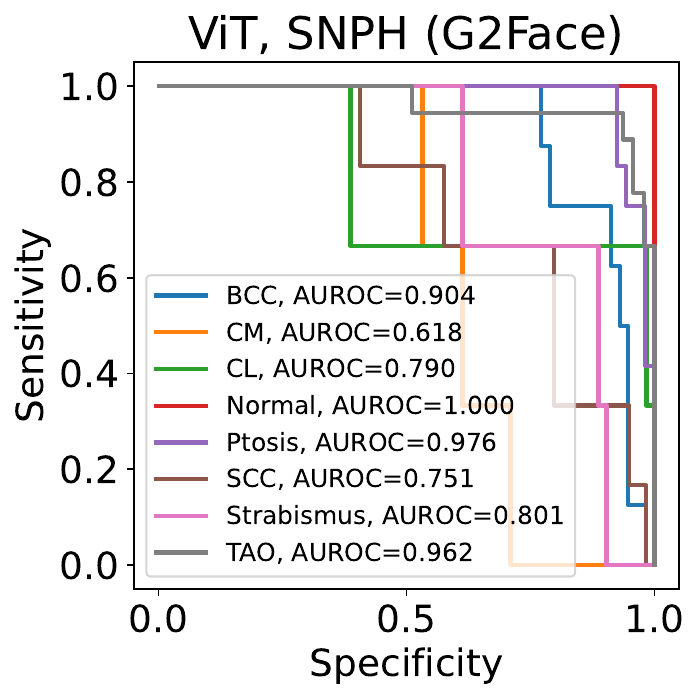}&
					\includegraphics[width=0.255 \linewidth]{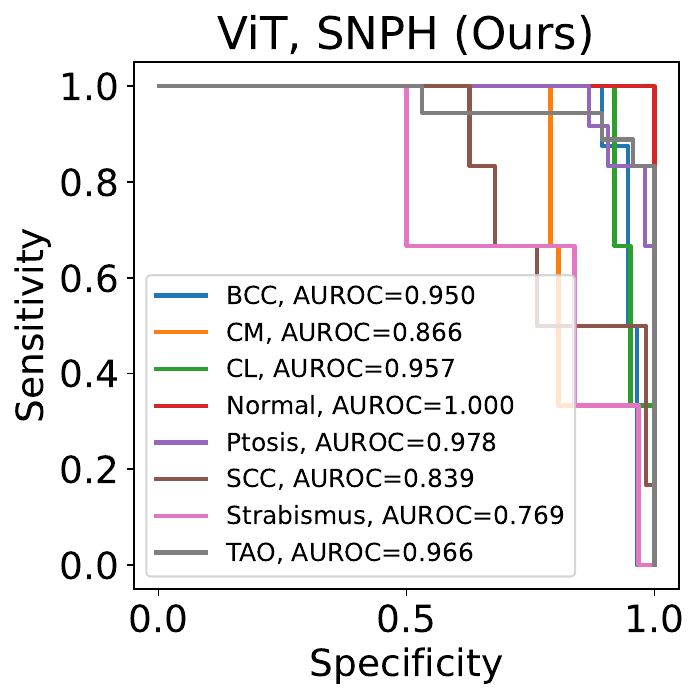}
				\end{tabular}
			\end{minipage}
			
		\end{minipage}
		
		\caption{
			\textbf{Compatibility to AI Diagnostic Models.}
			\textbf{a)} Diagnosis consistency between protected and original images, using ResNet50 diagnostic model.
			\textbf{b)} Diagnosis consistency, using ViT diagnostic model.
			\textbf{c)} Area Under the Receiver Operating Characteristic Curve (AUROC) comparison. Models were trained with five distinct random seeds and evaluated on the test set to produce five replicates. To assess whether ROFI's performance superiority over the second-best method is significant, we calculated the $P$ value using a two-sided t-test.
			\textbf{d)} ROC curves of the ViT diagnostic model on SNPH for images protected by different privacy protection methods. The ROC curves for the ViT diagnostic model on ECXHCSU and the ResNet50 diagnostic models on both SNPH and ECXHCSU are provided in the Supplementary Figure 7, Supplementary Figure 8 and Supplementary Figure 9.
		}
		\label{fig:trail_AI_model_auc}
	\end{figure*}
	
		\begin{figure*}[!thbp]

		\begin{minipage}	{1\linewidth}
			\begin{minipage}	{0.24\linewidth}
				\centering
				\setlength{\tabcolsep}{-1mm}
				\begin{tabular}{c}
					\multicolumn{1}{l}{\textsf{\small \textbf{a)}} } \\
					\includegraphics[width=1 \linewidth]{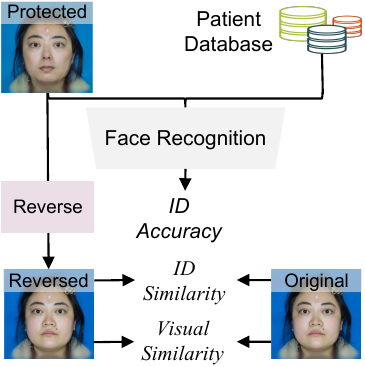}
				\end{tabular}
			\end{minipage}
			\begin{minipage}	{0.53\linewidth}
				\newcommand{\myarraystretch}{0.5}
				\centering
				\setlength{\tabcolsep}{-1mm}
				\begin{tabular}{cc}
					\multicolumn{2}{l}{\textsf{\small \textbf{b)}} } \\
					\includegraphics[width=0.50 \linewidth]{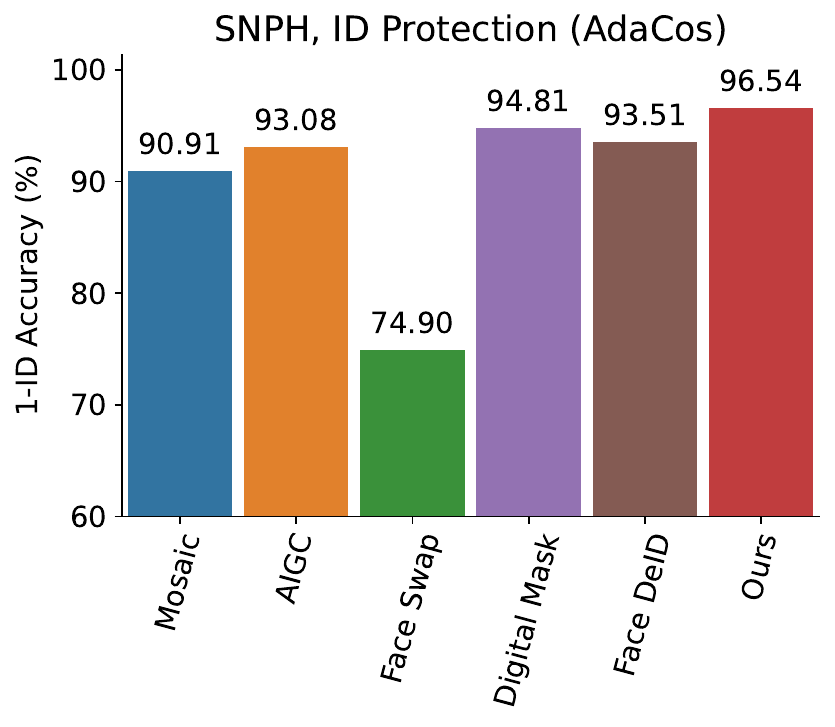} &
					~\includegraphics[width=0.50 \linewidth]{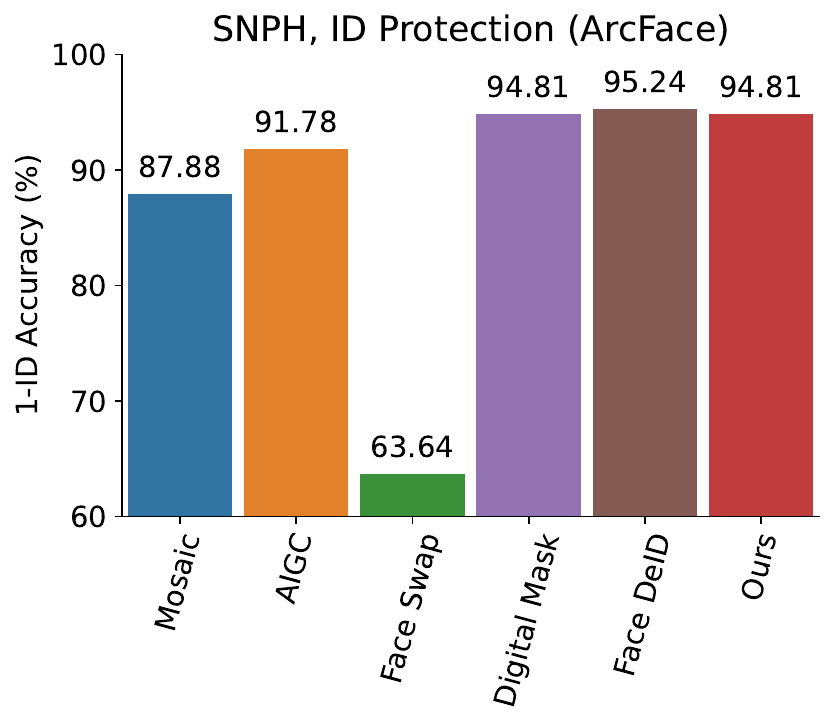}
					%				&
					%				~\includegraphics[width=0.33 \linewidth]{code/face_recognition_exp/plot_recognition_rate/face_id_rank1_human.pdf}
				\end{tabular}
			\end{minipage}
			\begin{minipage}	{0.18\linewidth}
				\newcommand{\myarraystretch}{0.5}
				\centering
				\setlength{\tabcolsep}{-1mm}
				\begin{tabular}{c}
					\multicolumn{1}{l}{\textsf{\small \textbf{c)}} } \\
					\includegraphics[width=1 \linewidth]{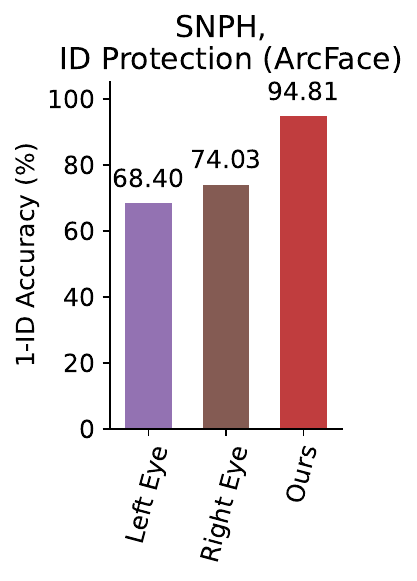} 
				\end{tabular}
			\end{minipage}
			\begin{minipage}	{0.99\linewidth}
				
				\begin{minipage}	{0.32\linewidth}
					\newcommand{\myarraystretch}{0.5}
					\centering
					\setlength{\tabcolsep}{-1mm}
					\begin{tabular}{cc}
						\multicolumn{2}{l}{\textsf{\small \textbf{d)}} } \\
						\includegraphics[width=0.49 \linewidth]{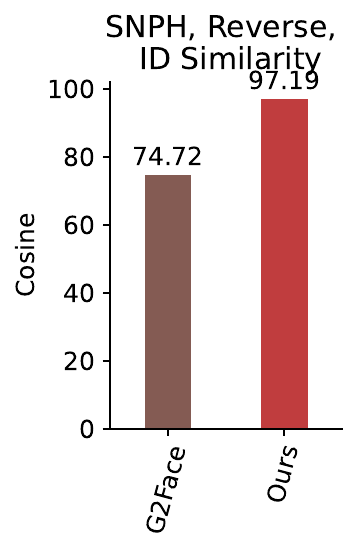} &
						~\includegraphics[width=0.49 \linewidth]{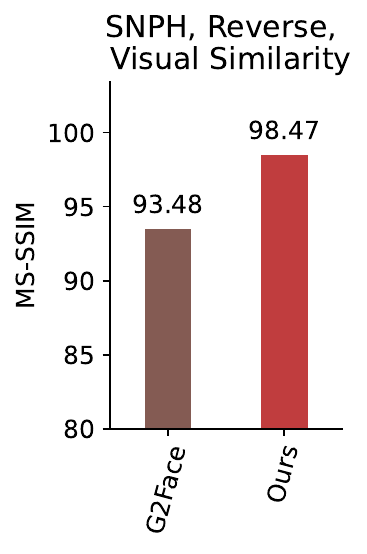}
					\end{tabular}
				\end{minipage}
				\hfill
				\begin{minipage}{0.215\linewidth}
					\vspace{-3mm}
					\centering
					\begin{tabular}{c}
						\multicolumn{1}{l}{\textsf{\small \textbf{e)}} } \\
						\vspace{-5mm}
						\includegraphics[width=1 \linewidth]{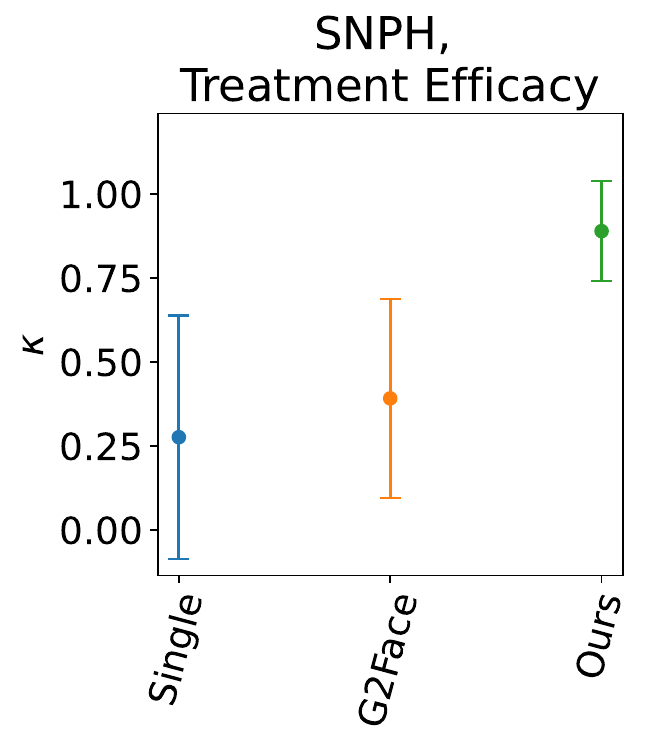} 
					\end{tabular}
				\end{minipage}
				\hfill
				\begin{minipage}{0.42\linewidth}
					\vspace{-3mm}
					\centering
					\begin{tabular}{c}
						\multicolumn{1}{l}{\textsf{\small \textbf{f)}} } \\
						\includegraphics[width=1 \linewidth]{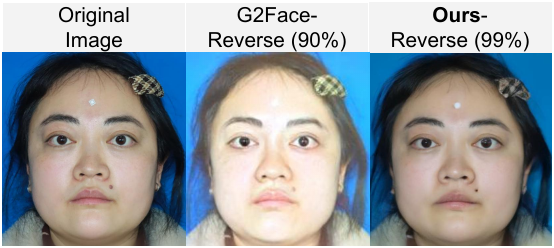} 
					\end{tabular}
				\end{minipage}
			\end{minipage}
		\end{minipage}
		\caption{
			\textbf{Face Protection and Reversible Capability.}
			\textbf{a)} Workflow for evaluating the privacy protection and the reversible reconstruction.
			\textbf{b)} ID protection rate. A higher value indicates a greater proportion of images successfully protected. We use two prominent face recognition methods, AdaCos and ArcFace.
			\textbf{c)} The conventional eye-cropping method is ineffective for privacy protection. Even a single cropped eye can lead to successful patient identification.
			\textbf{d)} Quantitative assessment of reversibly reconstructed image quality.
			\textbf{e)} TED hormone treatment efficacy, by comparing the current image with the retrieved history image.
			\textbf{f)} Qualitative comparison of the reconstructed images.
		}
		\label{fig:face_recognition}
		%	\vspace{-1cm}
	\end{figure*}

	\begin{figure*}[!htbp]
	\centering
	\setlength{\tabcolsep}{0mm}
	\begin{tabular}{c}
		\includegraphics[width=0.99\linewidth]{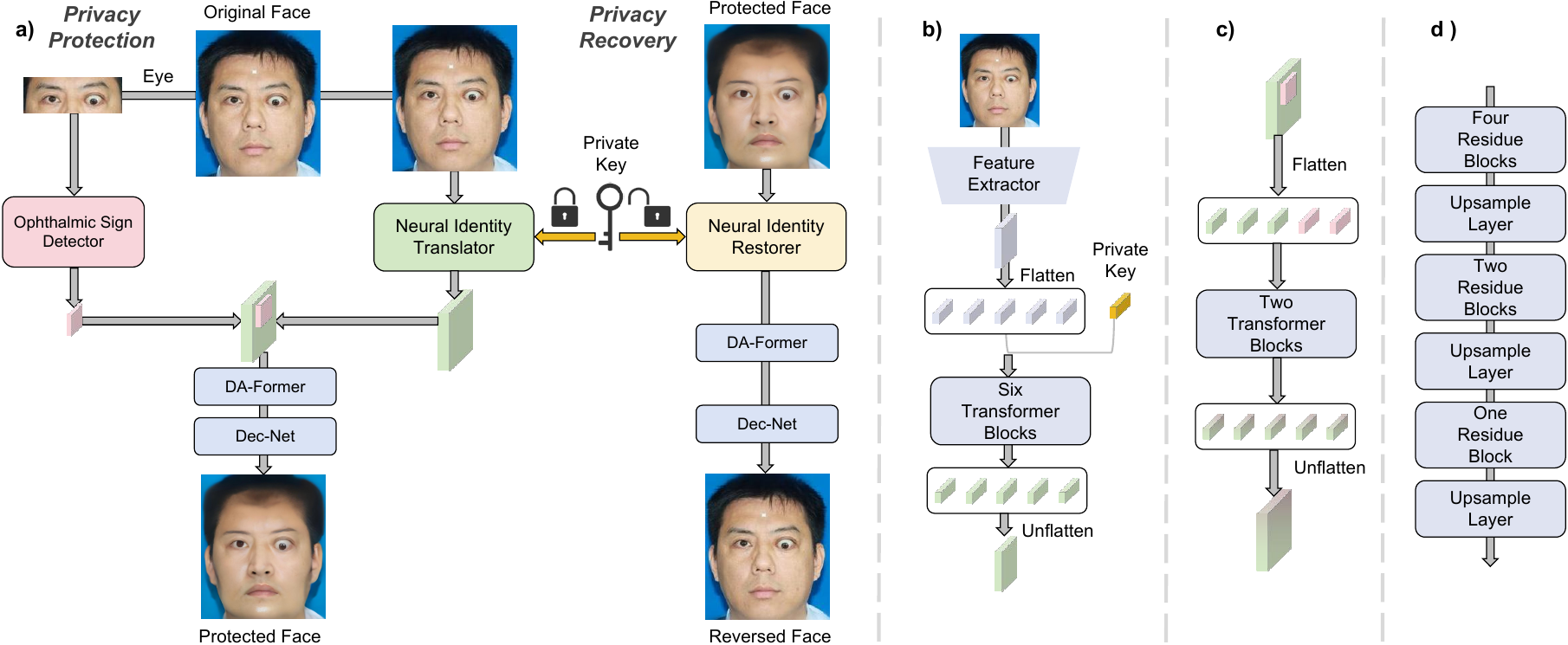} \\
		\vspace{-4mm} 
	\end{tabular}
	\caption{
		\textbf{Network Architectures of ROFI.}
		\textbf{a)} Overview of ROFI. All components within ROFI are deep neural networks, which are jointly trained with the learning objectives detailed in Section~\ref{sec:framework_objective}.
		\textbf{b)} Neural Identity Translator.
		\textbf{c)} DA-Former.
		\textbf{d)} Dec-Net.
		Flatten: reshape the 2D feature map of size $H \times W$ into 1D feature of size $HW$.
		Unflatten: reshape the 1D feature of size $HW$ to the 2D feature map of size $H \times W$.
		Icons are from https://uxwing.com/.
	}
	\label{fig:model_overview}
\end{figure*}

\begin{figure*}[!htbp]
	\centering
	\includegraphics[width=0.8 \linewidth]{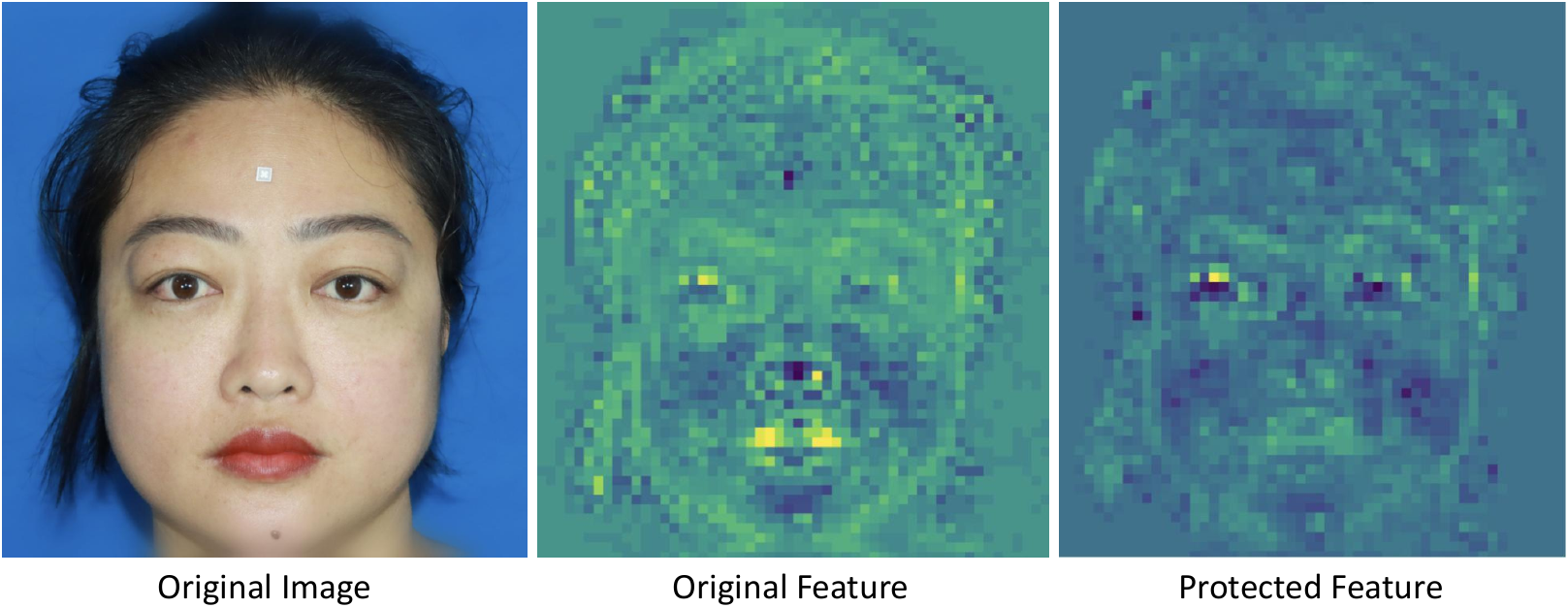}
	\caption{
		\textbf{Feature Visualization.}
		We visualize the facial feature maps before and after being protected by our ROFI model.
	}
	\label{fig:vis_feat}
\end{figure*}

\begin{figure}[!htbp]
	\centering
	\setlength{\tabcolsep}{0mm}
	\begin{tabular}{c}
		\includegraphics[width=0.9\linewidth]{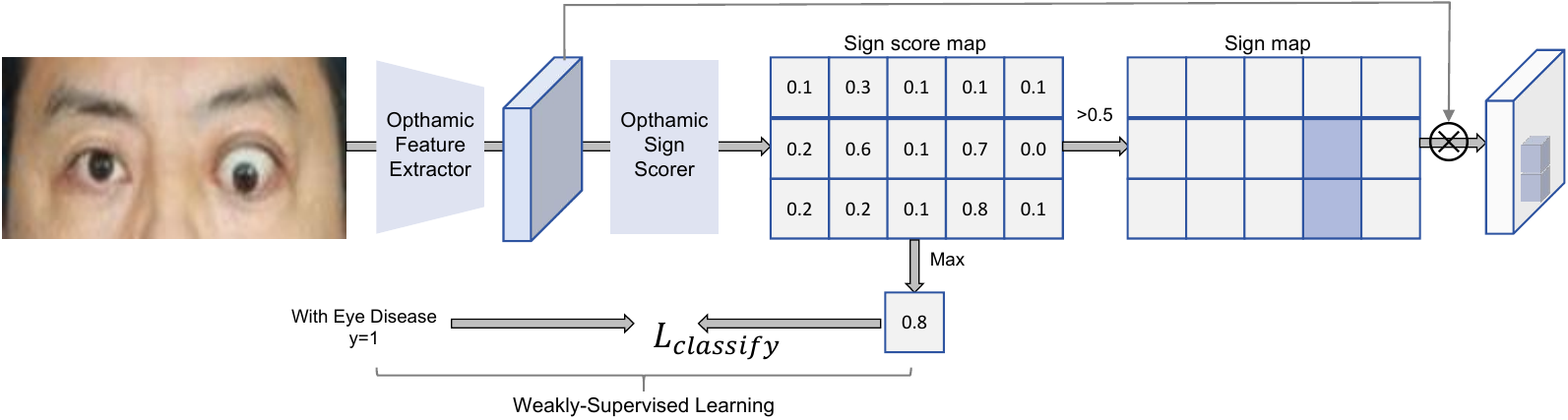} \\
		\vspace{-4mm} 
	\end{tabular}
	\caption{
		\textbf{Illustration of the Ophthalmic Sign Detector.}
		This detector learns to identify region-wise ophthalmic signs using a \textbf{``region-score-max''} strategy in a weakly-supervised manner, relying solely on image-level annotations, namely, the image is with eye disease or not.
		Features are masked with the sign map to preserve those essential for ophthalmic diagnosis, using the spatial-wise multiplication operation $\otimes$.
	}
	\label{fig:sign_weak_learn}
\end{figure}

	\begin{figure*}[!thbp]
	\centering
	\includegraphics[width=0.8 \linewidth]{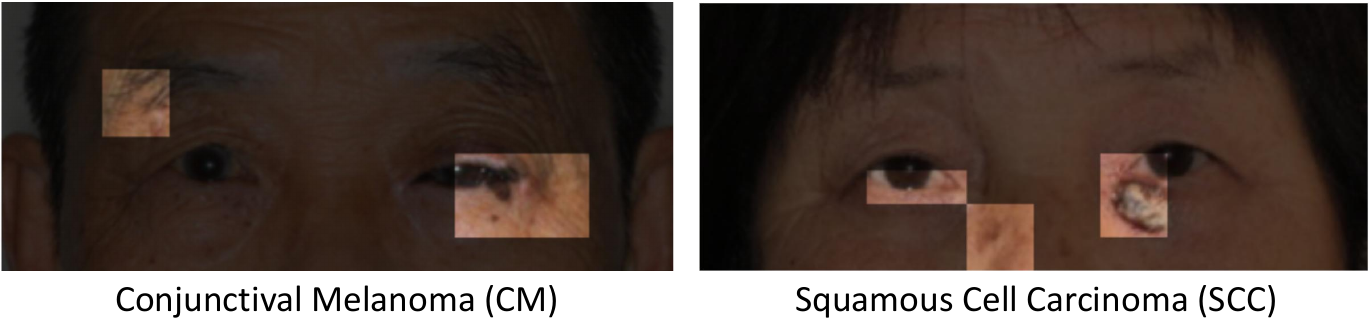}
	\caption{
		{\textbf{Sign Map Visualization.} Sign map is automatically detected by the ophthalmic sign detector.}
	}
	\label{fig:vis_mask}
\end{figure*}

\end{document}